\pdfoutput=1
\PassOptionsToPackage{table}{xcolor}

\documentclass[11pt]{article}

\usepackage[preprint]{acl}

\usepackage{times}
\usepackage{latexsym}
\usepackage{booktabs}
\usepackage{booktabs}
\usepackage{amssymb}
\usepackage{tabularx}
\usepackage{subcaption}
\usepackage{colortbl}
\usepackage[T1]{fontenc}
\usepackage[utf8]{inputenc}

\usepackage{microtype}

\usepackage{inconsolata}

\usepackage{graphicx}
\usepackage{amsmath} 

\newcommand{\model}{\textsc{Aigve-Macs}}
\newcommand{\data}{\textsc{Aigve-Bench 2}}

\title{\model{}: Unified Multi-Aspect Commenting and Scoring Model for AI-Generated Video Evaluation}

\author{Xiao Liu \qquad Jiawei Zhang \\
        IFM Lab, University of California, Davis\\
        \texttt{xioliu@ucdavis.edu, jiwzhang@ucdavis.edu}}

\begin{document}
\maketitle
\begin{abstract}
The rapid advancement of AI-generated video models has created a pressing need for robust and interpretable evaluation frameworks. Existing metrics are limited to producing numerical scores without explanatory comments, resulting in low interpretability and human evaluation alignment. To address those challenges, we introduce \model{}, a unified model for AI-Generated Video Evaluation(AIGVE), which can provide not only numerical scores but also multi-aspect language comment feedbacks in evaluating these generated videos. Central to our approach is \data{}, a large-scale benchmark comprising 2,500 AI-generated videos and 22,500 human-annotated detailed comments and numerical scores across nine critical evaluation aspects. Leveraging \data{}, \model{} incorporates recent Vision-Language Models with a novel token-wise weighted loss and a dynamic frame sampling strategy to better align with human evaluators. Comprehensive experiments across supervised and zero-shot benchmarks demonstrate that \model{} achieves state-of-the-art performance in both scoring correlation and comment quality, significantly outperforming prior baselines including GPT-4o and VideoScore. In addition, we further showcase a multi-agent refinement framework where feedback from \model{} drives iterative improvements in video generation, leading to $53.5\%$ quality enhancement. This work establishes a new paradigm for comprehensive, human-aligned evaluation of AI-generated videos. We release the \data{} and \model{} at \url{https://huggingface.co/xiaoliux/AIGVE-MACS}.

\end{abstract}

\section{Introduction}
\label{sec:introduction}
With the rapid advancement of video generation models such as Sora~\cite{sora2024}, HunyuanVideo~\cite{HunyuanVideo}, and Mochi-1~\cite{genmo2024mochi}, AI-generated videos are becoming increasingly photorealistic and temporally coherent, significantly narrowing the gap between synthetic and real-world visual content. Following provided textual instruction prompts, AI-generated videos are gaining widespread adoption across domains such as entertainment, advertising, education, and virtual reality, offering cost-effective solutions and enabling more personalized content experiences for both creators and consumers~\cite{vermacomparing, Zhang2024The}.

\begin{figure*}[t]
    \centering
    \includegraphics[width=0.92\linewidth]{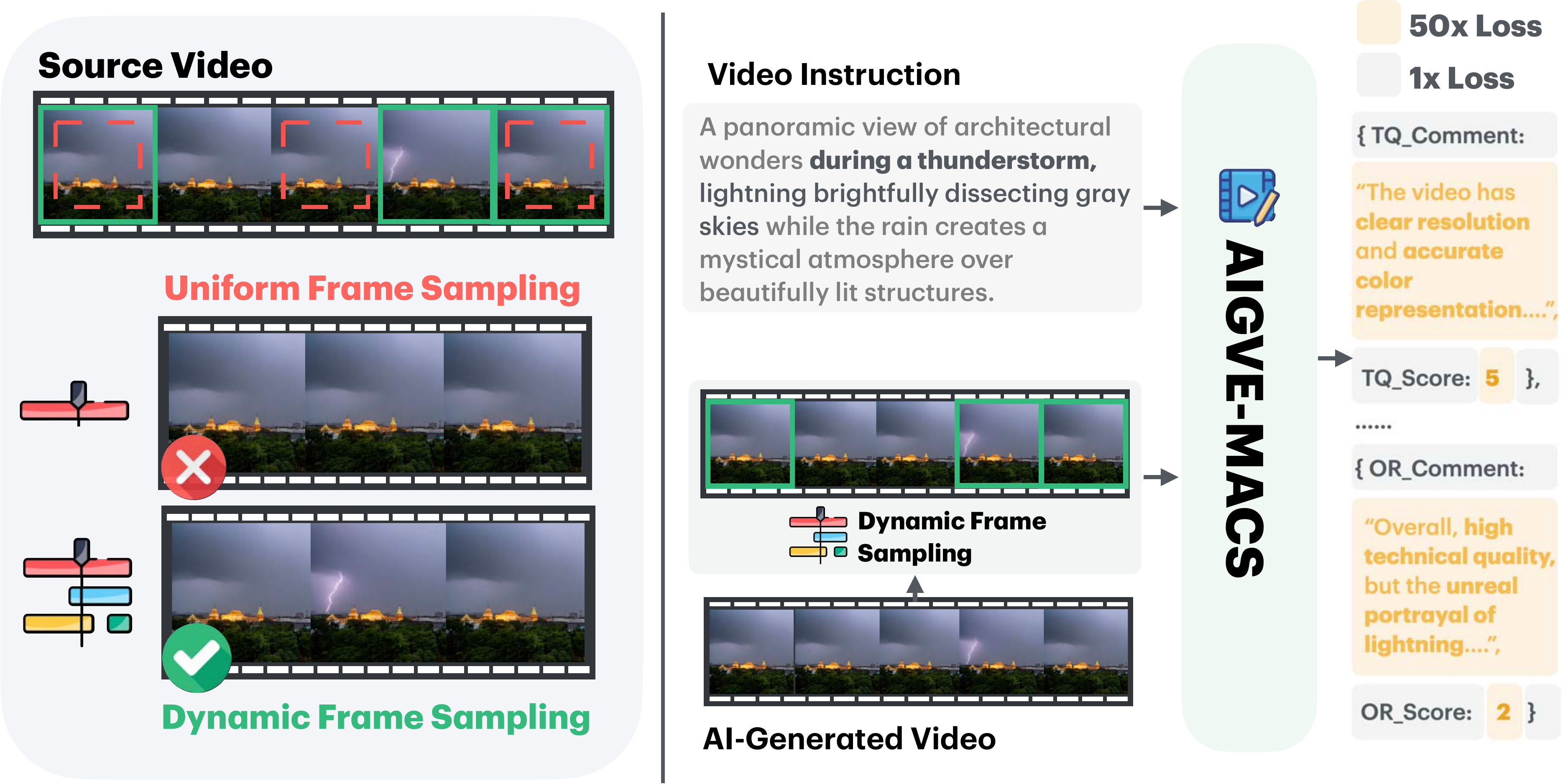}
    \caption{\model{} Pipeline. The left side of the diagram illustrates our Dynamic Frame Sampling strategy, which selects key moments based on content variation to better capture temporal dynamics. The right side highlights the Token-Wise Weighted Loss, which emphasizes score and comment tokens to improve alignment with human evaluations. TQ and OR refer to Technical Quality and Overall, respectively. Tokens in the yellow box are assigned a loss weight of 50, while those in the gray box are assigned a loss weight of 1.}
    \label{fig:pipline}
\end{figure*}

Despite significant progress, AI-generated videos still face persistent challenges such as limited spatial resolution, object distortions, and misalignment with user instructions~\cite{lee2024videorepair, wang2025swap,he2024videoscore}. These issues highlight the growing importance of rigorous evaluation, which plays a critical role in guiding the development and refinement of video generation models. However, research on evaluation methodologies has lagged behind the rapid advances in video generation itself, leaving a gap in systematic, interpretable, and comprehensive evaluation techniques~\cite{liu2024surveyaigve}.

Existing evaluation methods for AI-generated videos exhibit several key limitations. First, they often rely on outdated numerical metrics such as FVD~\cite{fvd} and IS~\cite{barratt2018note}, which are insufficient for capturing the nuanced qualities of modern AI-generated content. Effective AI-Generated Video Evaluation (AIGVE) must account not only for intrinsic video quality but also for alignment with user-provided instructions~\cite{xiang2025aigvetoolaigeneratedvideoevaluation, liu2024surveyaigve}. Therefore, evaluating video quality with one single scalar metric is inherently limiting. While recent work has made progress by decomposing quality into aspects such as technical quality, motion dynamics, and text-to-video alignment~\cite{he2024videoscore,chen2025finger,bansal2024videophy}, many approaches still rely on aggregating legacy metrics not originally designed for this task~\cite{huang_vbench_2023,liu_evalcrafter_2023}. These metrics often overlap in what they measure or fail to address key dimensions altogether, resulting in evaluation pipelines that are fragmented, difficult to interpret, and lacking comprehensive coverage.

Second, although recent studies have explored the use of unified Vision Language Models (VLMs) for evaluating AI-generated videos, they often fail to fully leverage the generative strengths of these models. This shortcoming stems largely from the difficulty of reliably prompting or finetuning VLMs to produce accurate, human-aligned evaluative outputs. Rather than generating natural language assessments directly, many existing approaches either attach a scoring head to predict numerical values from the VLM's hidden states, thereby bypassing its language generation capabilities~\cite{he-etal-2024-videoscore}, or reformulate evaluation as a series of binary yes/no questions~\cite{chen2025finger, zhang2024evaluation}. These strategies reduce the VLM to a classification tool, limiting its ability to provide nuanced, context-aware evaluations. Furthermore, the question-based approach relies heavily on handcrafted prompts, which struggle to capture the complexity and diversity of real-world video quality dimensions. Besides, due to the constrained context length of current VLMs, it is impossible to input the full token sequence of an entire video. Instead, existing methods typically sample a fixed number of frames uniformly across the video, which can overlook brief but critical dynamic changes.

Third, due to the absence of datasets that pair aspect-wise scores with explanatory comments, existing video evaluation approaches primarily focus on producing numerical scores~\cite{he-etal-2024-videoscore,kou_subjective-aligned_2024-1,miao2024t2vsafety,chen2024gaiarethinkingactionquality} while neglecting the generation of human-like comments that offer richer, more actionable feedback. However, we argue that such comments are actually essential for understanding the specific strengths and weaknesses of AI-generated videos. By providing detailed qualitative insights, explanatory comments can guide model improvement more effectively than scalar scores alone, which will also be demonstrated with showcases to be provided in this paper.

To address the aforementioned challenges, we first introduce and release the \data{}, a large-scale, human-annotated benchmark comprising 500 diverse prompts, 2,500 videos generated by five state-of-the-art video generation models, and 22,500 high-quality human score and comment annotations spanning nine carefully designed evaluation aspects. The prompts, videos, scores, and comments in \data{} are diligently
constructed and rigorously validated to ensure comprehensive coverage of diverse video generation scenarios, making the dataset both robust and representative of real-world applications.

Building on this benchmark, we propose to finetune Qwen2.5-VL-7B~\cite{bai2025qwen2} and present \model{}, a unified evaluation model that provides both numerical score and natural language comment feedback evaluations across the nine critical aspects as covered in the \data{}. The overall pipeline of the proposed \model{} is also illustrated in Figure~\ref{fig:pipline}. By introducing a dynamic frame sampling strategy 
as well as a weighted loss on both comment and score tokens, \model{} achieves accurate numerical predictions while fully leveraging the generative capabilities of VLMs to generate language comment to justify those scores. This joint modeling approach enhances the transparency, interpretability, and practical utility of automated video evaluation. Extensive experiments and ablation studies show that \model{} achieves state-of-the-art performance in both supervised and zero-shot settings on \data{} and multiple evaluation benchmarks, significantly outperforming existing methods in both numerical scoring and comment generation. 

What's more, to further demonstrate the practical utility of our model, we introduce a multi-agent iterative refinement framework that leverages the generated scores and comments to progressively improve video quality. Experimental results indicate that the quality of generated videos can be enhanced by 53.5\% through this iterative process. This framework underscores the real-world applicability of \model{}, illustrating how continuous feedback loops can drive meaningful improvements in AI-generated content. To the best of our knowledge, this is the first work to jointly produce aspect-wise scores and natural language comments for AI-generated videos, providing a more holistic evaluation framework that aligns closely with human judgments.

In summary, our contributions are as follows:
\begin{itemize}
\item We construct \data, the first large-scale AIGVE benchmark covering nine evaluation aspects, with 22,500 human-annotated numerical scores and explanatory comments for 2,500 videos generated by several SOTA video generation models.

\item We propose \model{}, the first unified AIGVE model that jointly produces aspect-wise scores and natural language comments, which can be trained with both dynamic frame sampling strategy and weighted loss terms on both score and comment tokens.
\item Besides \data{}, we have also conducted extensive experiments and ablation studies on several other evaluation benchmarks, demonstrating that \model{} achieves state-of-the-art performance in both supervised and zero-shot settings, with strong alignment to human judgments.
\item We explore practical applications of our model by introducing a multi-agent iterative refinement framework for video generation models specifically, which uses the generated scores and comments to iteratively improve the generated video quality.
\end{itemize}

\section{Related Works}
\label{sec:related}
\subsection{Diffusion-based Video Generation Models}

Diffusion models have emerged as the dominant approach in video generation, building on their remarkable success in text-to-image generation~\cite{ramesh2021zero, ding2021cogview, betker2023improving}. \citet{ho2022videodiffusionmodels} pioneered video diffusion by extending the standard image diffusion architecture to video data using a 3D U-Net architecture, while \citet{singer2022makeavideotexttovideogenerationtextvideo} improved computational efficiency through pseudo-3D convolutional layers. Subsequent works advanced the field: \citet{zhang2023show1marryingpixellatent} proposed a hybrid approach combining pixel-based and latent-based diffusion models, and \citet{lin2024ctrladapterefficientversatileframework} adapted ControlNets for enhanced control capabilities. Recent commercial models like OpenAI's Sora~\cite{sora2024} and Tencet's HunyuanVideo~\cite{HunyuanVideo} have demonstrated significant advancements in generating highly realistic videos. These models showcase the rapid progress in video diffusion models, though challenges remain in maintaining consistent visual quality and accurate alignment with human instructions.

\subsection{AI-Generated Video Evaluation}

The field of AI-Generated Video Evaluation (AIGVE) is in its early stages and includes even more challenges than video quality assessment. Previous research proposes that evaluating AI-generated videos requires alignment with both human perception and instructions to ensure high-quality video generation that meets creators' intentions and viewers' expectations~\cite{liu2024surveyaigve}. Alignment with human perception focuses on evaluating video quality through traditional metrics like resolution and clarity, while ensuring consistency with physical world properties such as realistic textures and adherence to physical laws. Meanwhile, alignment with human instructions emphasizes how well videos mirror the detailed scenarios, actions, and narratives specified in text descriptions, ensuring the generated content fulfills creators' creative and communicative objectives. These aspects have evolved separately, with perception alignment progressing from statistical approaches~\cite{fvd, barratt2018note} to advanced architectures like DOVER~\cite{wu_exploring_2023-1} for aesthetic assessment and BVQI~\cite{BVQI} for CLIP-based evaluation, while instruction alignment developed methods like TIFA~\cite{hu2023tifa} and CLIPScore~\cite{hessel-etal-2021-clipscore} for evaluating text-visual consistency.

Recent comprehensive frameworks attempt to unify these aspects. VBench~\cite{huang_vbench_2023} evaluates across 16 dimensions, and EvalCrafter~\cite{liu_evalcrafter_2023} assesses four key aspects including visual quality and text-video alignment. VIDEOSCORE~\cite{he-etal-2024-videoscore} leverages large language models for holistic assessment. However, these methods either rely on collections of traditional metrics or provide single scores without interpretable feedback, highlighting the need for more sophisticated evaluation approaches.
\section{\data{}}
\label{sec:data}
% add figure of comment length distribution
Our \data{} extends the previous benchmark dataset \textsc{AIGVE-Bench}~\cite{xiang2025aigvetoolaigeneratedvideoevaluation}, which contains multi-aspect numerical scores for AI-generated videos across nine evaluation dimensions. We introduce a rigorous comment process pipeline to enrich \textsc{AIGVE-Bench} with human-written, aspect-specific justifications. This pipeline ensures that each textual explanation is directly aligned with the corresponding multi-aspect numerical evaluation scores, offering a more comprehensive and interpretable understanding of video quality. The remainder of this section will first present he key characteristics of of \textsc{AIGVE-Bench} and then detail our comment processing pipeline.

\subsection{Previous Benchmark: \textsc{AIGVE-Bench}}
\textsc{AIGVE-Bench}~\cite{xiang2025aigvetoolaigeneratedvideoevaluation} is a comprehensive AIGVE benchmark dataset contains 2,500 AI-generated videos created from 500 diverse prompts, covering a broad range of real-world scenes and interactions. Each prompt falls into one of two categories: \textit{global view}, which focuses on environmental and wide-angle scenes with natural dynamics such as weather changes, camera motion, and \textit{close shot}, which captures fine-grained interactions between humans, animals, plants, or objects. Videos are generated by five state-of-the-art models: Pyramid~\cite{jin2024pyramidal}, CogVideoX~\cite{yang2024cogvideox}, Genmo~\cite{genmo2024mochi}, Hunyuan~\cite{HunyuanVideo}, and Sora~\cite{sora2024}, with varying resolutions and frame rates, and are standardized to a uniform duration of 5 seconds to ensure sufficient temporal depth and comparability.

Each video is evaluated by expert raters across nine distinct evaluation aspects: Technical Quality, Dynamics, Consistency, Physics, Element Presence, Element Quality, Action/Interaction Presence, Action/Interaction Quality, and Overall. The detailed description of each aspect can be found in Appendix~\ref{app:data_details}, Table~\ref{tab:metrics}.

\subsection{Comment Processing Pipeline}
\label{sec:comment_pipeline}
\begin{figure}[t]
    \centering
    \includegraphics[width=1\linewidth]{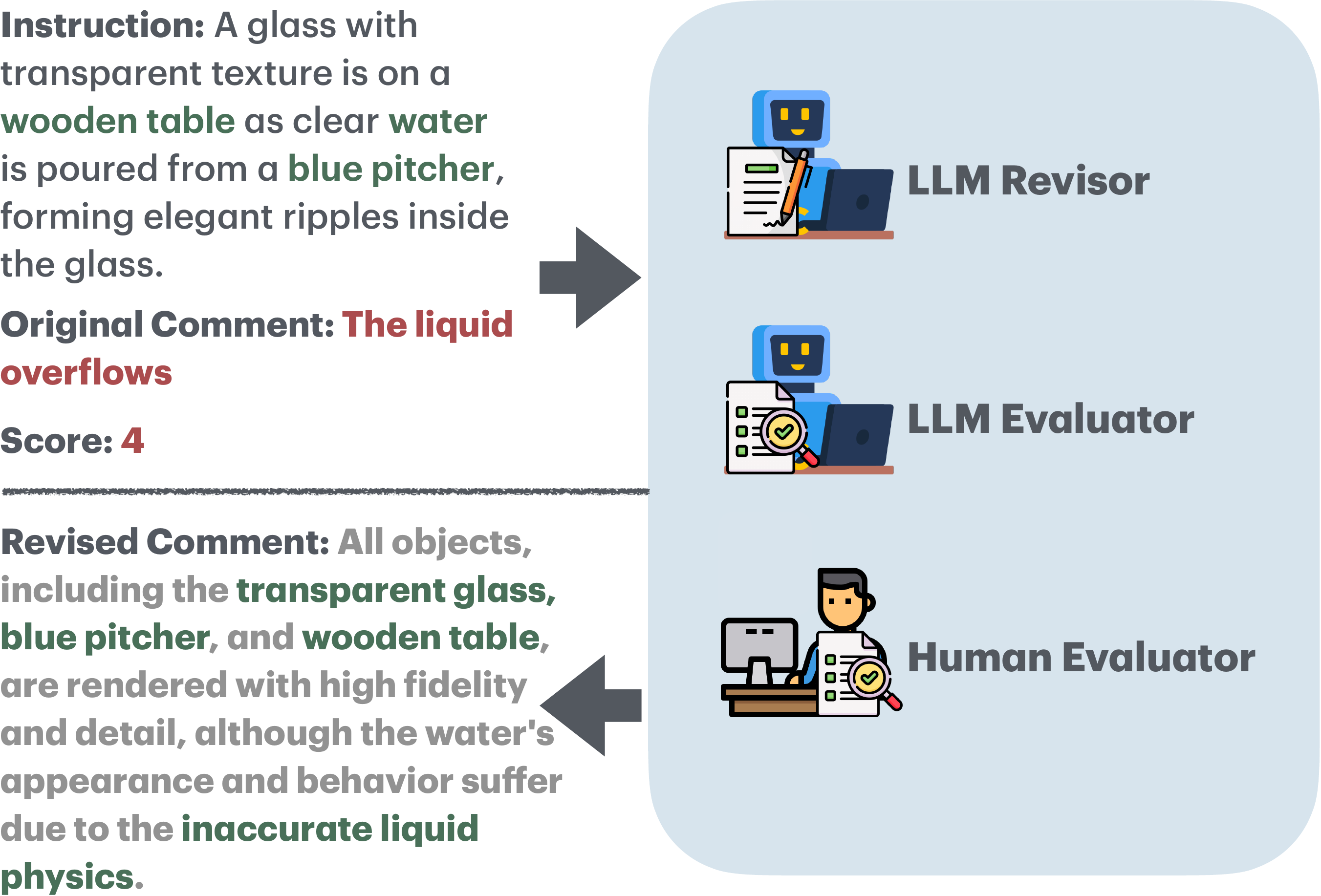}
    \caption{Overview of the comment processing pipeline for \data{}. The pipeline includes a revisor, validator, and human evaluator to ensure high-quality, consistent, and objective comments.}
    \label{fig:comment_pipeline}
\end{figure}

Although justification comments are collected in parallel with the scoring process, they present several challenges that limit their direct usability. First, since the comments are written by multiple annotators, they exhibit inconsistencies in writing style, terminology, and level of detail, introducing noise and increasing the complexity of downstream model training and evaluation. Second, some comments may not be fully aligned with the corresponding scores. As shown in the left side of Figure~\ref{fig:comment_pipeline}, when a video only has minor issues, annotators may focus solely on noting the issue without acknowledging the video's strengths. This can result in comments that fail to adequately justify a high score, reducing their effectiveness as training signals or explanatory references.

To address these challenges, we design a robust comment processing pipeline that revises the original comments to ensure consistency, objectivity, and alignment with the corresponding scores. As shown in Figure~\ref{fig:comment_pipeline}, we first employ a Large Language Model (LLM) as a revisor to refine and extend the original comments, conditioned on the video instructions and associated scores. The LLM is prompted to produce comments with a consistent and objective writing style, providing a comprehensive explanation that covers both the strengths and weaknesses relevant to the evaluation aspect.

Next, the revised comment is passed to a second LLM, which we refer to as the evaluator. This model checks whether the revised comment introduces any content that are not presented in the original comment or instruction, mitigating the risk of hallucination. Finally, a human evaluator manually reviews and, if necessary, rewrites the revised comments to ensure clarity, factual accuracy, and faithful reflection of the corresponding scores. Through this meticulous process, we extend each aspect-level score in \textsc{AIGVE-Bench} with a high-quality justification comment, resulting in a total of 22,500 comments with an average length of 267 words. The distribution of comment lengths is shown in Appendix~\ref{app:data_details}, Figure~\ref{fig:comment_length_distribution}.

\section{\model{}}
\label{sec:method}

To build the \model{} framework, we propose to finetune the VLMs with our \data{}, enabling VLMs to generate human-aligned evaluation scores and explanatory comments. Specifically, given an AI-generated video, we provide the model with the video instruction and video content. The VLM is trained to generate a JSON-formatted output containing both the aspect-specific score and the corresponding natural language comment.

Formally, given an AI-generated video $v$, the video's instruction $i$, the VLM is trained to generate a structured output:

\begin{equation}
y = \operatorname{VLM}(i, v|\theta),
\end{equation}

where $y = \{C, S\}$ denotes the JSON-formatted output containing aspect-specific evaluation comments $C$ and scores $S$, $\theta$ is the learnable parameter of the VLM.

However, as discussed in Section~\ref{sec:introduction}, existing methods are facing two major challenges. First, they struggle to effectively leverage the native language generation capabilities of VLMs to produce human-aligned evaluation scores and comments. Second, the commonly adopted uniform frame sampling strategy may overlook brief yet critical dynamics, leading to incorrect evaluation. To address these issues, we first introduce a token-wise weighted loss in Section~\ref{sec:twwl}, which encourages the model to focus more on generating informative comments and accurate scores by highlighting these tokens during training. Additionally, in Section~\ref{sec:dfs}, we propose a dynamic frame sampling strategy to better capture key moments in the video.

\begin{figure}[t]
    \centering
    \includegraphics[width=0.6\linewidth]{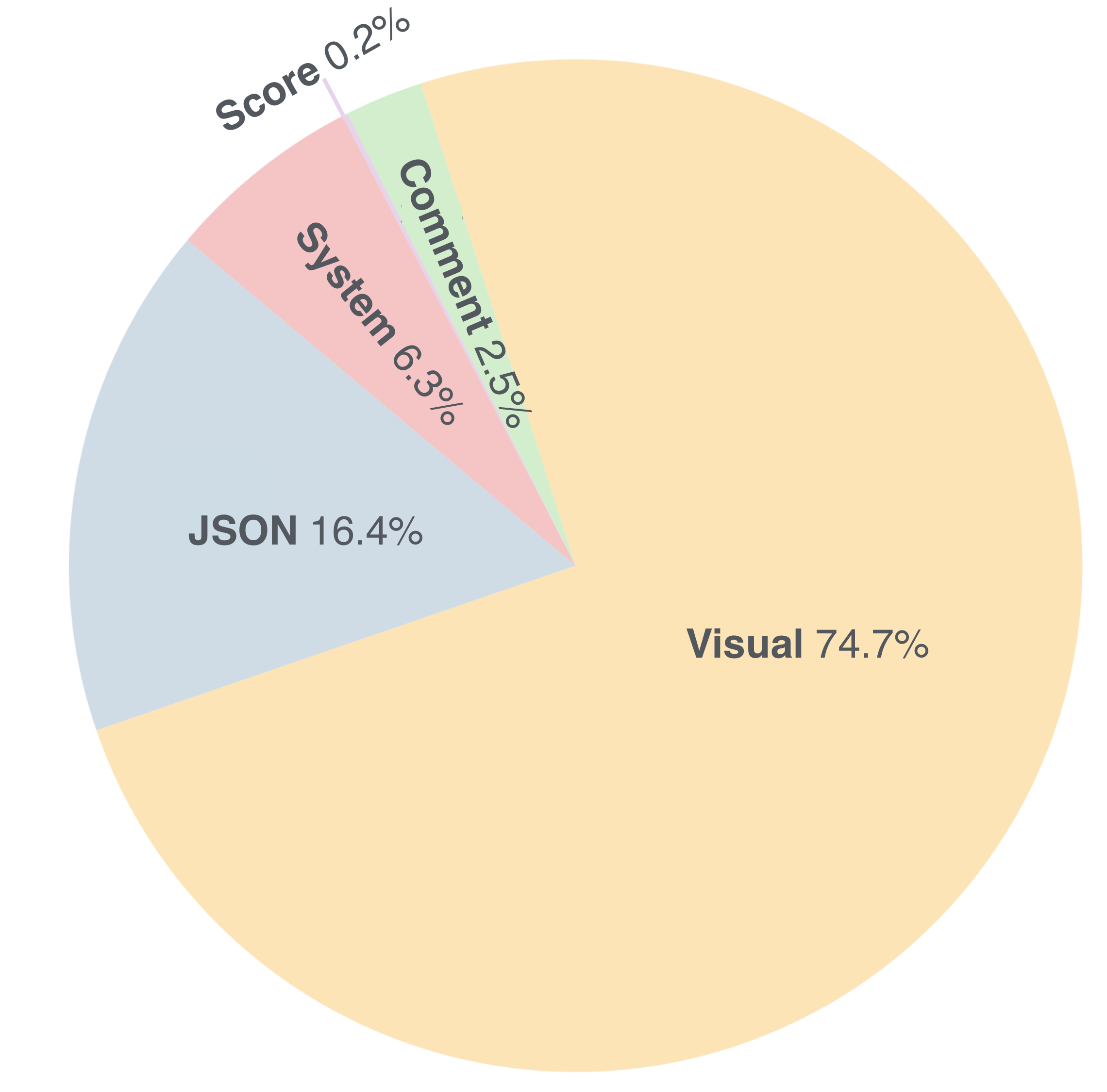}
    \caption{Averaged ratios of different types of input tokens in the input sequence. The five types of tokens are: (1) system prompt tokens, (2) visual tokens, (3) JSON structure tokens, (4) comment tokens, and (5) numerical score tokens.}
    \label{fig:input_tokens}
\end{figure}

\subsection{Token-wise Weighted Loss}
\label{sec:twwl}
To better understand the challenge of leveraging language generation ability, we analyze the input tokens used by VLMs in our task. Specifically, there are five types of input tokens: (1) system prompt tokens that describe the task, (2) visual tokens that encode the video content, (3) JSON structure tokens that maintain the output format, (4) comment tokens that contain user comments, and (5) score tokens that indicate the evaluation scores. Figure~\ref{fig:input_tokens} illustrates the averaged ratios of each type of input token in the input. We observe that the system prompt tokens, viusal tokens, and JSON structure tokens, which are not directly related to the task, occupy a significant portion of the input, while the task-related comment tokens and score tokens are relatively sparse. This unbalanced distribution of input tokens will lead to the VLMs focusing more on the system prompt, visual content, and output format, while neglecting the user comments and scores.

To address this issue, we propose a simple yet effective token-wise weighted loss that encourages the model to focus on accuratly generate comment and score tokens. During tokenization, we identify comment and score tokens and assign them a higher weight $\alpha$ in the loss function. The weighted loss is defined as follows:

\begin{equation}
\mathcal{L} = -\sum_{t=1}^{T} w_t \cdot \log p(y_t \mid y_{<t}, i, v),
\end{equation}

where $T$ is the total number of tokens in the output sequence, $y_t$ is the target token at position $t$, $x$ is the input sequence, and $p(y_t \mid y_{<t}, i, v)$ is the predicted probability of token $y_t$. The token weight $w_t$ is defined as:

\begin{equation}
w_t =
\left\{
\begin{array}{ll}
\alpha, & \text{if } y_t \text{ is a comment or score token}, \\
1, & \text{otherwise}.
\end{array}
\right.
\end{equation}

Besides, inspired by reasoning schema, we put the comment content prior to the numerical score in training data to encouage the model to build stronger connections between the comment and the score to enhance the comment-score alignment. The process is shown on the right part of Figure~\ref{fig:pipline}.

\subsection{Dynamic Frame Sampling}
\label{sec:dfs}
To avoid missing the key moments in the video, we propose a dynamic frame sampling strategy that selects frames based on content variation instead of uniform sampling. As shown at the bottom-left of Figure~\ref{fig:pipline}, this approach allows the evaluation model to focus on the most informative frames, leading to more accurate and aspect-aware assessments.

To select representative frames based on content changes, we compute a frame-wise difference score as follows:

\begin{equation}
\footnotesize
\Delta(f^t, f^{t-1}) = \frac{1}{H \times W} 
\sum_{i=1}^{H} \sum_{j=1}^{W} 
\mathbb{I}\left( \left| f^t_{i, j} - f^{t-1}_{i, j} \right| > 0 \right)
\end{equation}

where \( f^t \in \mathbb{R}^{H \times W} \) denotes the grayscale frame with width $W$ and height $H$ at time \( t \), and \( \mathbb{I}(\cdot) \) is the indicator function. A frame \( f^t \) is selected if it satisfies both:
\begin{equation}
\Delta(f^t, f^{t-1}) > \theta \quad \text{and} \quad t - t_{\text{last}} \geq \gamma,
\end{equation}
where \( \theta \) is a content change threshold and \( \gamma \) is the minimum frame gap to prevent redundant selection. If the number of selected frames exceeds the target number \( N \), we uniformly subsample them. If no frames meet the criteria, we fallback to uniform sampling. The implementation details of \model{} is presented in Appendix~\ref{app:imple}.

\section{Experiment}
\subsection{Benchmark Datasets}

To comprehensively evaluate the performance of \model{}, we conduct experiments on four benchmark datasets: one supervised dataset, \data{}-\textsc{Test}, and three zero-shot datasets: VIDEOFEEDBACK-Test~\cite{he-etal-2024-videoscore}, GenAI-Bench~\cite{li2024genaibenchevaluatingimprovingcompositional}, and VBench~\cite{huang_vbench_2023}. For fair comparison, we use the same test samples as those used in VideoScore~\cite{he-etal-2024-videoscore} for the zero-shot settings. The details of each dataset are as follows:

\noindent\paragraph{\data{}-\textsc{Test}}
We perform stratified sampling to select 200 examples from the test split of \data{}, ensuring a balanced distribution across the \textit{global-view} and \textit{close-shot} categories, and comprehensive coverage of all motion types and object categories. This approach preserves evaluation efficiency while maintaining diversity and representativeness.

For score evaluation, we compute Spearman's rank correlation coefficient ($\rho$) between the predicted scores and ground-truth scores across all nine evaluation aspects. To assess the quality of generated comments, we adopt a suite of metrics: ROUGE-1 and ROUGE-L~\cite{lin-2004-rouge} for information coverage, UniEval-Fact~\cite{zhong2022unifiedmultidimensionalevaluatortext} for faithfulness, BERTScore~\cite{zhang2020bertscoreevaluatingtextgeneration} for semantic similarity, and G-Eval~\cite{liu2023gevalnlgevaluationusing} to assess overall comment quality.

\noindent\paragraph{VideoFeedback-Test}
VideoFeedback~\cite{he-etal-2024-videoscore} decomposes AI-generated video quality into five evaluation aspects: Visual Quality, Temporal Consistency, Dynamic Degree, Text Alignment, and Factual Consistency. We report Spearman’s $\rho$ between predicted and human-annotated scores for each aspect.

\begin{table*}[!t]
\centering
\resizebox{\textwidth}{!}{
\begin{tabular}{l|ccccccccc|c}
\toprule
\textbf{Method} & \textbf{Technical} & \textbf{Dynamic} & \textbf{Consistency} & \textbf{Physics} & \textbf{Element Pre} & \textbf{Element Qu} & \textbf{Act Pre} & \textbf{Act Qu} & \textbf{Overall} & \textbf{AVG} \\
\midrule
Random \ & -6.71 & 4.28 & -3.55 & -1.62 & -3.03 & 1.42 & -5.16 & 1.44 & -3.19 & -1.79 \\
\midrule
\multicolumn{11}{c}{\textbf{Feature-based Metrics}} \\
\midrule
CLIP-sim & 9.12 & 6.54 & 5.79 & -4.45 & 19.34 & 7.93 & 23.85 & 8.25 & 21.58 & 10.88 \\
BLIP-sim & 12.02 & 10.92 & 12.24 & 3.69 & 22.80 & 9.42 & 17.02 & 10.34 & 19.09 & 13.06 \\
CLIP-temp & 16.46 & 4.26 & \underline{27.69} & \underline{23.51} & 11.07 & 25.12 & -2.62 & 20.20 & 16.63 & 15.81 \\
PickScore & 22.48 & 5.90 & 11.36 & 6.55 & 26.25 & 17.69 & 20.37 & 13.24 & 24.28 & 16.46 \\
\midrule
\multicolumn{11}{c}{\textbf{Modeling-based Metrics}} \\
\midrule
LightVQA+ & -3.68 & -7.58 & -8.08 & -10.73 & -0.16 & -6.33 & 4.77 & -7.40 & -4.70 & -4.88 \\
GSTVQA & 17.92 & 13.40 & 15.97 & 1.23 & -2.65 & 15.91 & 9.81 & 9.68 & 20.71 & 11.33 \\
SimpleVQA & 24.50 & 11.50 & 16.58 & 0.28 & 2.41 & 18.30 & 7.76 & 3.58 & 21.22 & 11.79 \\
\midrule
\multicolumn{11}{c}{\textbf{VLM-based Metrics}} \\
\midrule
Qwen2.5-VL & 8.77 & 4.00 & 1.24 & -6.01 & 9.19 & 10.19 & 18.74 & 0.72 & 9.59 & 6.27 \\
VideoLLaMA3 & 15.94 & \underline{19.44} & 11.70 & 13.21 & -3.13 & 12.27 & 13.61 & -0.69 & 11.58 & 10.44 \\
GPT-4.1 & \underline{36.49} & 5.81 & 26.68 & 19.87 & \underline{27.22} & 28.77 & 32.75 & 20.22 & 29.98 & 25.31 \\
GPT-4o & 34.71 & 7.05 & 18.12 & 20.28 & 23.10 & \underline{30.47} & \underline{36.57} & \underline{31.58} & \underline{38.57} & \underline{26.72} \\
\midrule
Videoscore & -9.50 & -8.20 & -0.20 & 20.10 & 9.70 & -7.50 & -3.10 & -0.60 & -7.30 & -0.73 \\
VideoPhy & 0.10 & 4.00 & -1.40 & -5.60 & 0.80 & -1.30 & 11.90 & 2.70 & 9.70 & 2.32 \\
DSGScore & 6.92 & -7.62 & 10.62 & 11.71 & 7.62 & 9.63 & 1.04 & 8.91 & 3.47 & 5.81 \\
VIEScore & 15.61 & 3.39 & 2.27 & 6.22 & 9.99 & 8.19 & 8.47 & 8.22 & 11.98 & 8.26 \\
TIFA & 17.81 & 8.83 & 9.62 & 3.85 & 16.67 & 12.41 & 17.90 & 5.87 & 17.78 & 12.30 \\
\midrule
\model{} & \textbf{40.60} & \textbf{57.31} & \textbf{61.49} & \textbf{64.36} & \textbf{40.32} & \textbf{40.81} & \textbf{44.31} & \textbf{60.71} & \textbf{59.88} & \textbf{52.20} \\
\bottomrule
\end{tabular}
}
\caption{Scoring Correlation Evaluation Result on \data{}. The underlined score represents the best zero-shot model. Element Pre and Element Qu represents Element Presence and Quality. Act Pre and Qu represents Action Presence and Quality.}
\label{tab:video_eval_metrics}
\end{table*}

\begin{table}[!t]
\centering
\small
\scalebox{0.97}{
\begin{tabular}{l|cc|c|c|c}
\toprule
\textbf{} & \textbf{R1} & \textbf{RL} & \textbf{BS} & \textbf{UF} & \textbf{GE} \\
\midrule
GPT-4o & 18.30 & 15.86 & 74.90 & 40.84 & 2.10 \\
GPT-4.1 & 15.80 & 12.94 & 73.99 & 43.99 & 2.10 \\
QWen2.5-VL & 17.95 & 15.31 & 74.31 & 42.32 & 2.37 \\
VideoLLama & 19.99 & 17.67 & 75.35 & 40.21 & 2.18 \\
\midrule
\model{} & \textbf{49.50} & \textbf{38.00} & \textbf{85.87} & \textbf{57.04} & \textbf{3.42} \\
\bottomrule
\end{tabular}
}
\caption{Evaluation Results for Comments. R1, RL, BS, UF, and GE represent ROUGE-1, ROUGE-L, BERTcore, UniEval-Fact, and G-Eval respectively.}
\label{tab:comment_eval}
\end{table}

\noindent\paragraph{GenAI-Bench}
GenAI-Bench~\cite{li2024genaibenchevaluatingimprovingcompositional} is derived from GenAI-Arena, a human preference dataset where users compare pairs of AI-generated videos. We use \model{} to predict per-video scores across the nine evaluation aspects, then infer pairwise preferences by comparing average scores. Following the VideoScore protocol, we report pairwise accuracy against human-labeled preferences.

\noindent\paragraph{VBench}
VBench~\cite{huang_vbench_2023} evaluates AI-generated videos using a collection of existing automatic metrics. Following VideoScore, we calculate pairwise accuracy across five aspects: Technical Quality, Subject Consistency, Dynamic Degree, Motion Smoothness, and Overall Consistency.

\subsection{Baseline Methods}

To benchmark the performance of \model{}, we compare it against a comprehensive suite of existing video evaluation methods, organized into three main categories:

\paragraph{Feature-based Metrics}
These methods evaluate videos by extracting handcrafted or pretrained model features. This category includes traditional quality metrics such as PIQE~\cite{venkatanath2015blind} and BRISQUE~\cite{mittal2012no}, perceptual similarity-based scores like SSIM-sim~\cite{1284395}, SSIM-dyn~\cite{1284395}, MSE(Mean Square Error)-dyn, and DINO-sim~\cite{caron2021emerging}, as well as text-video alignment metrics such as CLIP-sim~\cite{kou2024subjective}, CLIP-temp~\cite{liu_evalcrafter_2023}, CLIP-Score~\cite{hessel-etal-2021-clipscore}, X-CLIP-Score~\cite{ma2022x}, BLIP-sim~\cite{li2022blipbootstrappinglanguageimagepretraining}, and PickScore~\cite{Kirstain2023PickaPicAO}.

\begin{table*}[t]
\centering
\resizebox{0.95\textwidth}{!}{
\begin{tabular}{l|ccccc|c}
\toprule
\textbf{Method} & \textbf{Visual Quality} & \textbf{Temporal} & \textbf{Dynamic Degree} & \textbf{Text Alignment} & \textbf{Factual} & \textbf{Average} \\
\midrule
Random & -3.1 & 0.5 & 0.4 & 1.1 & 2.9 & 0.4 \\
VideoScore & 86.2 & 80.3 & 77.6 & 59.4 & 82.1 & 77.1 \\
\midrule
\multicolumn{7}{c}{\textbf{Feature-based Metrics}} \\
\midrule
PIQE & -17.7 & -14.5 & 1.2 & -3.4 & -16.0 & -10.1 \\
BRISQUE & -32.4 & -26.4 & -4.9 & -8.6 & -29.1 & -20.3 \\
CLIP-sim & 21.7 & 29.1 & -34.4 & 2.0 & 26.1 & 8.9 \\
DINO-sim & 19.4 & 29.6 & -37.9 & 2.2 & 24.0 & 7.5 \\
SSIM-sim & 33.0 & \underline{30.6} & -31.3 & 4.7 & 30.2 & 13.4 \\
MSE-dyn & -20.3 & -24.7 & \underline{38.0} & 3.3 & -23.9 & -5.5 \\
SSIM-dyn & -31.4 & -29.1 & 31.5 & -5.3 & -30.0 & -12.9 \\
CLIP-Score & -10.9 & -10.0 & -14.7 & -0.3 & -0.3 & -7.2 \\
X-CLIP-Score & -3.2 & -2.7 & -7.3 & 5.9 & -2.0 & -1.9 \\
\midrule
\multicolumn{7}{c}{\textbf{VLM-based Metrics}} \\
\midrule
LLaVA-1.5 & 9.4 & 8.0 & -2.2 & 11.4 & 15.8 & 8.5 \\
LLaVA-1.6 & -8.0 & -4.1 & -5.7 & 1.4 & 0.8 & -3.1 \\
Idefics2 & 4.2 & 4.5 & 8.9 & 10.3 & 4.6 & 6.5 \\
Gemini-1.5-Flash & 24.1 & 5.0 & 20.9 & 21.3 & \underline{32.9} & 20.8 \\
Gemini-1.5-Pro & \underline{35.2} & -17.2 & 18.2 & \underline{26.7} & 21.6 & 16.9 \\
GPT-4o & 13.6 & 17.6 & 28.2 & 25.7 & 30.2 & 23.0 \\
\midrule
\model{} (Ours) & \textbf{30.8} & \textbf{27.5} & \textbf{18.4} & \textbf{24.4} & \textbf{21.7} & \underline{\textbf{24.6}} \\
\bottomrule
\end{tabular}
}
\caption{Evaluation Result on VideoFeedback-Test. The underlined score represents the best zero-shot model.}
\label{tab:video_eval_summary}
\end{table*}

\begin{table*}[!ht]
\centering
\resizebox{0.95\textwidth}{!}{
\begin{tabular}{l|c|cccccc}
\toprule
\textbf{Benchmark} & \textbf{GenAI-Bench} & \textbf{Technical Quality} & \textbf{Subject Consistency} & \textbf{Dynamic Degree} & \textbf{Motion Smoothness} & \textbf{Overall Consistency} & \textbf{Average} \\
\midrule
Random & 37.7 & 44.5 & 42.0 & 37.3 & 40.5 & 44.8 & 41.82 \\
\midrule
\multicolumn{8}{c}{\textbf{Feature-based Metrics}} \\
\midrule
PIQE & 34.5 & 60.8 & 44.3 & 71.0 & 45.3 & 53.8 & 55.04 \\
BRISQUE & 38.5 & 56.7 & 41.2 & 75.5 & 41.2 & 54.2 & 53.76 \\
CLIP-sim & 34.1 & 47.8 & 46.0 & 34.8 & 44.7 & 44.2 & 43.50 \\
DINO-sim & 31.4 & 49.5 & 51.2 & 24.7 & 55.5 & 41.7 & 44.52 \\
SSIM-sim & 28.4 & 30.7 & 46.2 & 24.5 & 54.2 & 27.2 & 36.56 \\
MSE-dyn & 34.2 & 32.8 & 31.7 & 81.7 & 31.2 & 39.2 & 43.32 \\
SSIM-dyn & 38.5 & 37.5 & 36.3 & \underline{84.2} & 34.7 & 44.5 & 47.44 \\
CLIP-Score & 45.0 & 57.8 & 46.3 & 71.3 & 47.0 & 52.2 & 54.92 \\
X-CLIP-Score & 41.4 & 44.0 & 38.0 & 51.0 & 28.7 & 39.0 & 40.14 \\
\midrule
\multicolumn{8}{c}{\textbf{VLM-based Metrics}} \\
\midrule
LLaVA-1.5 & 49.9 & 42.7 & 42.3 & 63.8 & 41.3 & 8.8 & 39.78 \\
LLaVA-1.6 & 44.5 & 38.7 & 26.8 & 56.5 & 28.5 & 43.2 & 38.74 \\
Idefics1 & 34.6 & 20.7 & 22.7 & 54.0 & 27.3 & 33.7 & 31.68 \\
Gemini-1.5-Flash & 67.1 & 52.3 & 49.2 & 64.5 & 45.5 & 49.9 & 52.28 \\
Gemini-1.5-Pro & 60.9 & 56.7 & 43.3 & 65.2 & 43.0 & 56.3 & 52.90 \\
GPT-4o & 52.0 & 59.3 & 49.3 & 46.8 & 42.0 & 60.8 & 51.64 \\
VideoScore & 59.0 & 64.2 & 57.7 & 55.5 & 54.3 & \underline{61.5} & 58.64 \\
\midrule
\model{} (Ours) & \underline{\textbf{68.5}} & \underline{\textbf{65.5}} & \underline{\textbf{59.0}} & \textbf{74.5} & \underline{\textbf{55.8}} & \textbf{55.6} & \underline{\textbf{62.08}} \\
\bottomrule
\end{tabular}
}
\caption{Evaluation results on GenAI-Bench and VBench. The underlined score represents the best zero-shot model.}
\label{tab:genai_vbench_results}
\end{table*}

\paragraph{Modeling-based Metrics}
This group includes supervised deep learning models trained specifically to predict video quality. We include LightVQA+~\cite{Light_VQA_plus}, GSTVQA~\cite{9452150}, and SimpleVQA~\cite{cheng2025simplevqamultimodalfactualityevaluation}, which are adapted for multi-aspect video assessment.

\paragraph{VLM-based Metrics}
These approaches leverage VLMs to generate scores or conduct evaluations via either prompting or finetuning. Baselines include foundation models such as Qwen2.5-VL~\cite{bai_qwen-vl_2023}, VideoLLaMA3-7B~\cite{damonlpsg2025videollama3}, GPT-4.1~\cite{gpt41}, GPT-4o~\cite{gpt4o}, LLaVA-1.5-7B~\cite{liu2024improvedbaselinesvisualinstruction}, LLaVA-1.6-7B~\cite{liu2024llavanext}, Gemini-1.5-Flash~\cite{geminiteam2024gemini15unlockingmultimodal}, Gemini-1.5-Pro~\cite{geminiteam2024gemini15unlockingmultimodal}, Idefics1~\cite{laurençon2023obelicsopenwebscalefiltered}, and Idefics2~\cite{laurençon2024mattersbuildingvisionlanguagemodels}. In addition, we consider VLM-based evaluation frameworks like VideoScore~\cite{he-etal-2024-videoscore}, DSG Score~\cite{Cho2024DSG}, VIEScore~\cite{ku_viescore_2023}, and TIFA~\cite{hu2023tifa}.

\begin{figure*}[!t]
    \centering
    \includegraphics[width=0.95\linewidth]{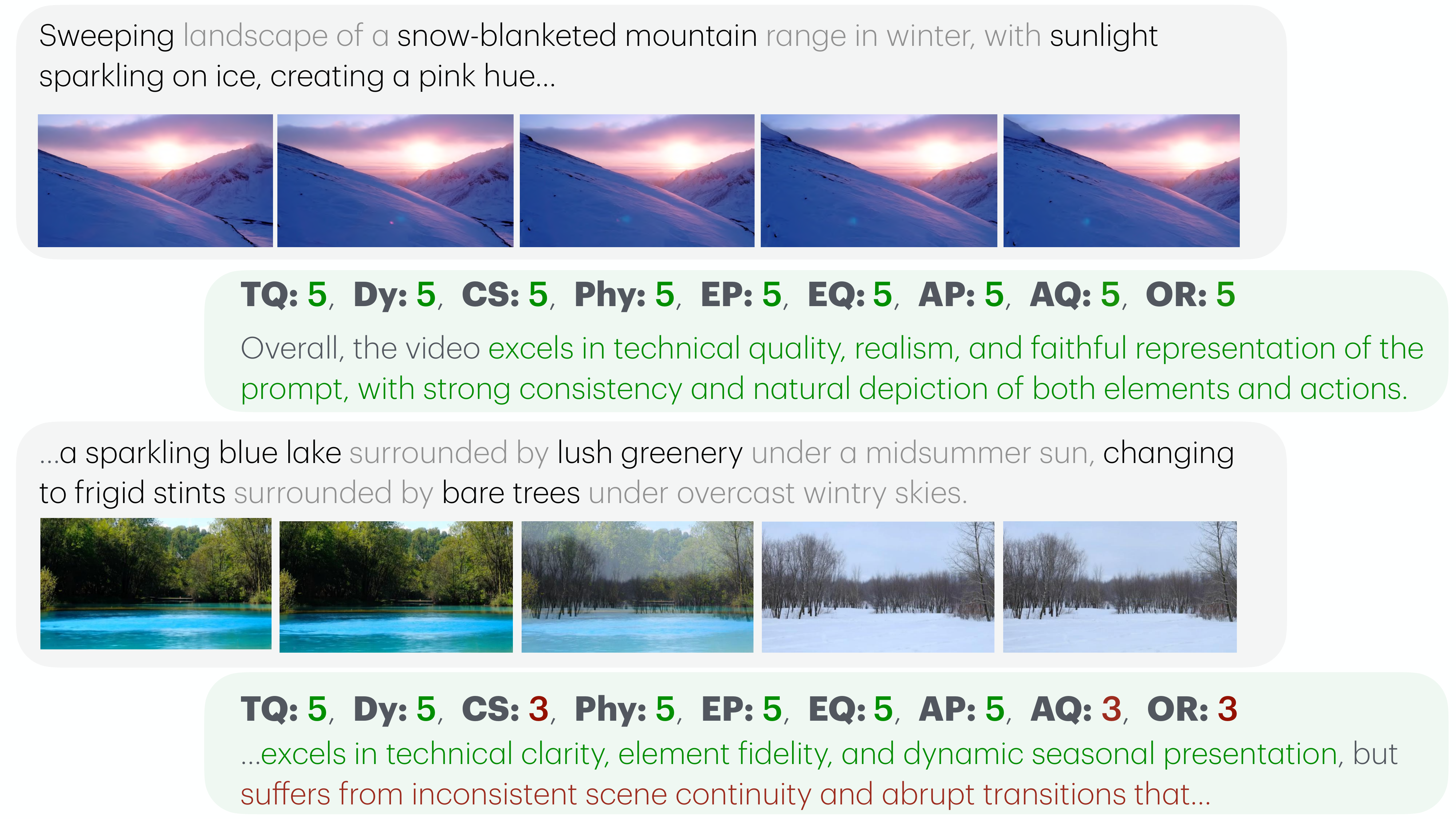}
    \caption{Case Study of \model{}. This example showcases \model{}'s ability to produce human-aligned evaluations and distinguish fine-grained quality differences in AI-generated videos. The model not only accurately assesses natural object fidelity and realism but also captures temporal dynamics, such as smooth versus abrupt scene transitions, leading to coherent and interpretable multi-aspect judgments.}
    \label{fig:case_study}
\end{figure*}

\subsection{Results}
\label{sec:results}
\textbf{\model{} achieves state-of-the-art alignment with human judgments, significantly outperforming all existing baselines in both score prediction and comment generation.} Table~\ref{tab:video_eval_metrics} and Table~\ref{tab:comment_eval} summarize \model{}'s performance on \data{} compared to a broad range of evaluation baselines. \model{} achieves the highest correlation across all nine evaluation aspects, with an average Spearman’s $\rho$ of 52.20\%, which nearly doubling the performance of the strongest VLM baseline, GPT-4o, and achieving up to 61\% improvement over its pretrained backbone, Qwen2.5-VL-7B~\cite{bai2025qwen2}.

In terms of comment generation, as shown in Table~\ref{tab:comment_eval}, \model{} also consistently outperforms all baselines across standard automatic metrics. These results highlight a fundamental limitation of current VLMs, which struggle to generate reliable, human-aligned evaluations. In contrast, \model{} effectively learns to generate both accurate scores and rich, faithful commentary that aligns closely with human judgment.

\noindent\textbf{\model{} establishes new state-of-the-art performance across all three benchmarks, demonstrating strong zero-shot generalization capabilities for diverse video evaluation tasks.} Table~\ref{tab:video_eval_summary} and Table~\ref{tab:genai_vbench_results} report zero-shot evaluation results on VideoFeedback~\cite{he2024videoscore}, GenAI-Bench~\cite{li2024genaibenchevaluatingimprovingcompositional}, and VBench~\cite{huang_vbench_2023}.

On GenAI-Bench, \model{} achieves the highest video preference accuracy, outperforming strong commercial VLMs such as Gemini-1.5-Pro and GPT-4o. On VBench, \model{} surpasses VideoScore by 4.24\%, while using only one-tenth of the training data.

Notably, on the VideoFeedback dataset, \model{} also achieves significant improvements over all open-source and commercial baselines, emerging as the best zero-shot model. By comparing the result in Table~\ref{tab:video_eval_metrics} and Table~\ref{tab:video_eval_summary}, while \model{} generalizes well to VideoFeedback without training on it, VideoScore, which is trained on VideoFeedback, fails to generalize to our \data{}-\textsc{Test}. This contrast highlights \model{}'s more robust and transferable evaluation capabilities that are not tied to the idiosyncrasies of any single dataset.

The superior supervised and zero-shot performance of \model{} validates the effectiveness of our finetuning strategy, which equips VLMs with the ability to evaluate AI-generated videos using both structured scores and natural language comments. It also underscores the high quality and broad coverage of \data{}, enabling the training of generalizable video evaluators that can seamlessly transfer to other benchmarks. To further validate our finetuning approach, we conduct an ablation study on the \data{}-\textsc{Test} set, as shown in Table~\ref{tab:ablation} in Appendix~\ref{app:ablation}. The results demonstrate that both token-wise weighted loss and dynamic frame sampling are crucial for achieving optimal performance. We present case study in Figure~\ref{fig:case_study}. More cases could be found in Appendix~\ref{app:case_studies}.

\section{Multi-Agent Iterative Refinement Framework}
\label{sec:multi-agent}

Our \model{} opens new possibilities for leveraging evaluation scores and comments to guide and optimize the video generation process. To initiate this exploration, we propose a multi-agent iterative refinement framework designed to progressively enhance video quality through feedback-driven revisions.

\begin{figure}[t]
    \centering
    \includegraphics[width=\linewidth]{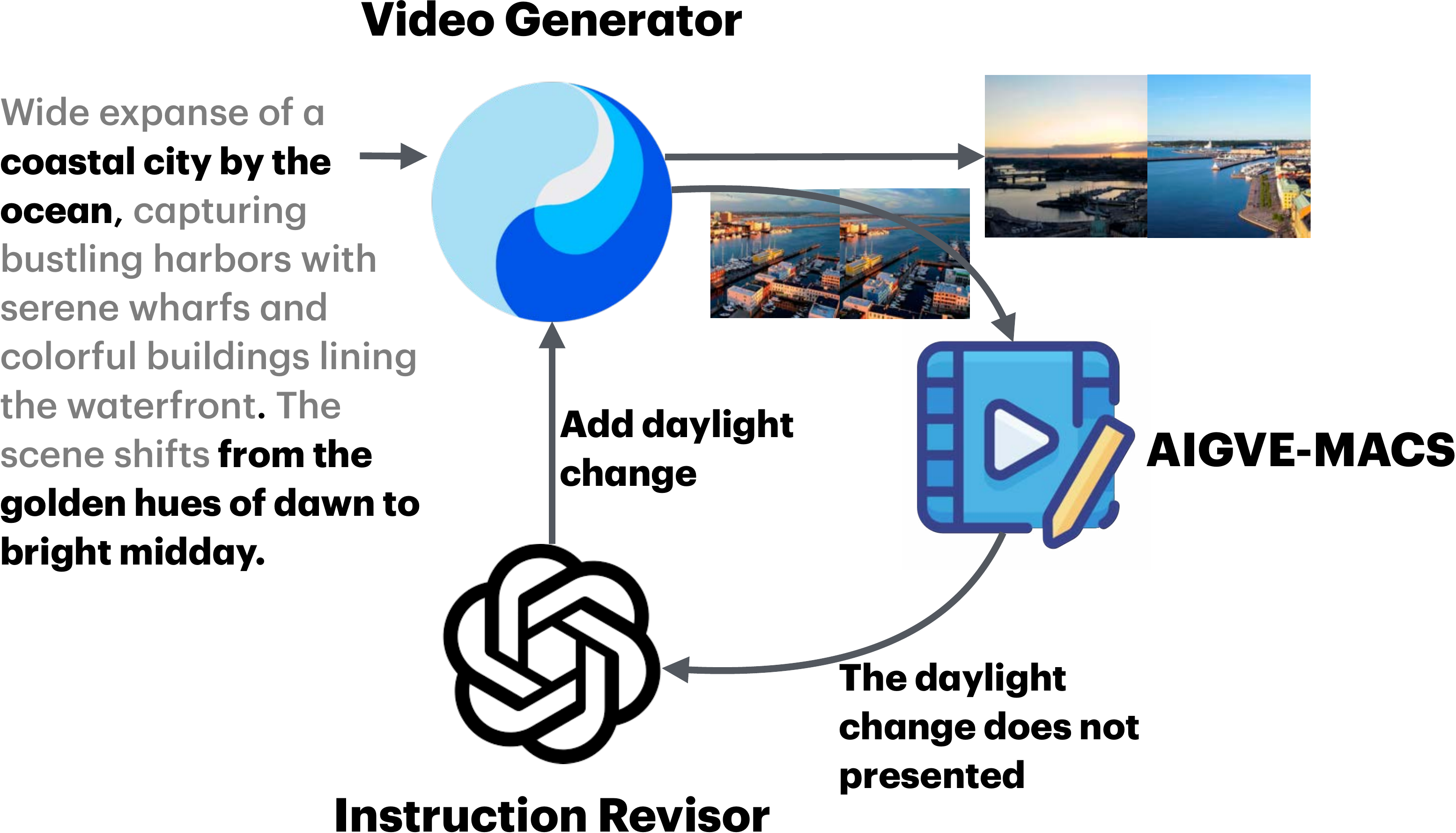}
    \caption{The pipeline of the multi-agent iterative refinement framework.}
    \label{fig:multi_agent}
\end{figure}

As shown in Figure~\ref{fig:multi_agent}, our framework consists of three main components: a Video Generator, an Instruction Revisor, and \model{}. In each iteration, the Video Generator produces a video based on the current user instruction, and then \model{} evaluates the generated video and provides feedback in the form of scores and comments. The Instruction Revisor refines the instruction based on the feedback, aiming to clarify or adjust the requirements for the next iteration. This process continues until the video reaches a satisfactory quality level or a maximum of 4 iterations.

To validate the effectiveness of our multi-agent framework, we use HunyuanVideo~\cite{HunyuanVideo} as the Video Generator, GPT-4.1~\cite{gpt41} as the Instruction Revisor, and \model{} as the evaluator. We conduct experiments on a subset of 50 low-quality videos (overall score < 3) sampled from \data{}-\textsc{Test}. The iteration continues until the overall score exceeds 4 or the iteration limit of 4 is reached.

\begin{figure}[t]
    \centering
    \includegraphics[width=0.87\linewidth]{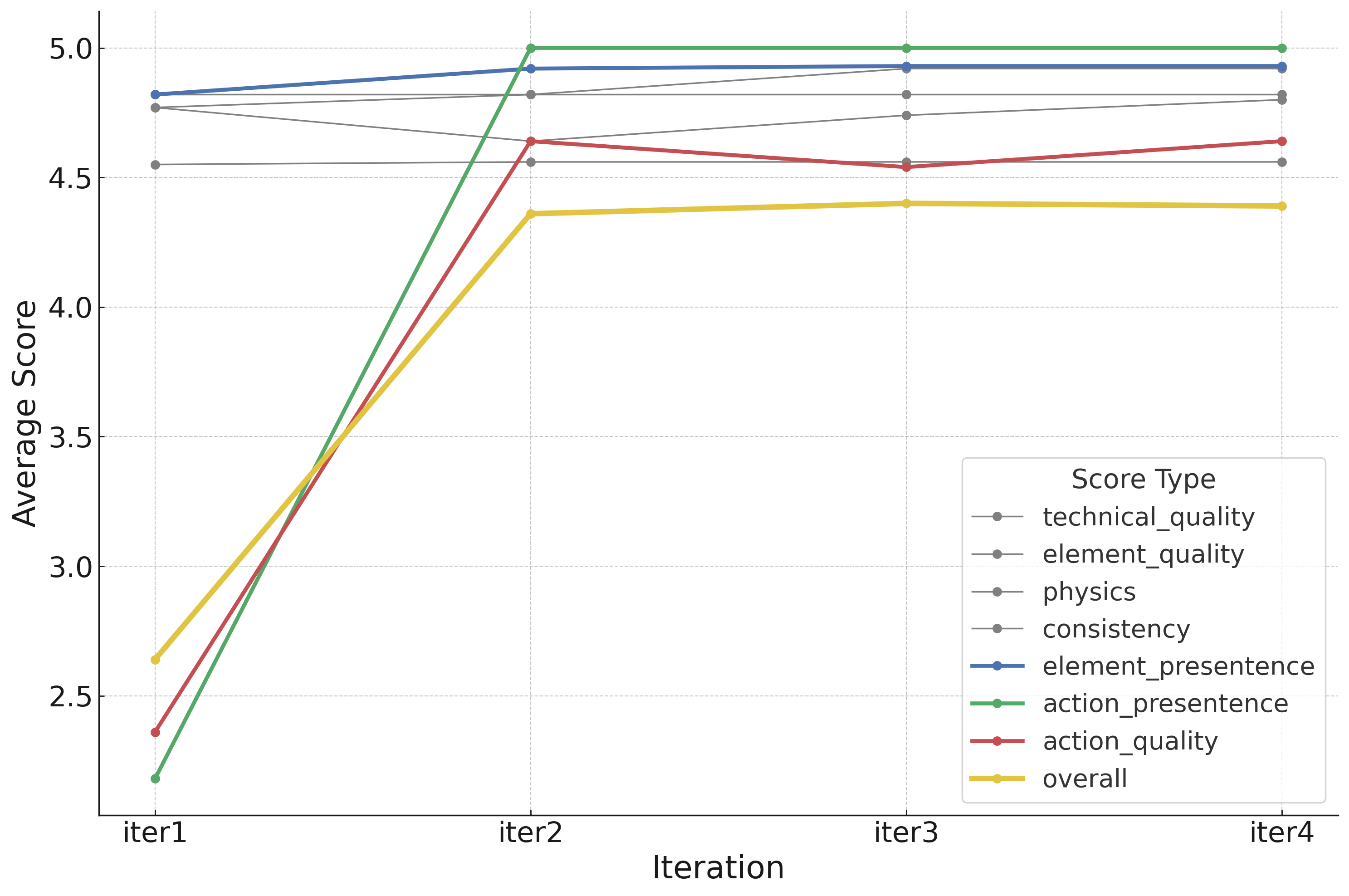}
    \caption{The result of the multi-agent iterative refinement framework.}
    \label{fig:multi_agent_res}
\end{figure}

Figure~\ref{fig:multi_agent_res} shows the average scores across refinement iterations. Notably, aspects related to  \textit{overall} score and instruction alignment such as element presence, action presence, and action quality, demonstrate significant improvements up to 53\%, particularly within the first two iterations.

In contrast, visual aspects like technical quality, element quality, and physics show only minor variation, as they are primarily constrained by the fixed video generator. This suggests that while our framework effectively improves instruction-driven dimensions, visual quality remains bounded by generator capacity.

\section{Conclusion}
We propose \model{}, a unified and interpretable evaluation framework that jointly predicts aspect-wise scores and explanatory comments for AI-generated videos. Trained on our human-annotated \data{} dataset, the model achieves state-of-the-art performance in both supervised and zero-shot settings, outperforming strong baselines in score correlation and comment quality across multiple benchmarks. Beyond evaluation, we demonstrate the utility of AIGVE-SCORE in a multi-agent refinement framework, showing its ability to drive meaningful improvements in video generation quality through feedback-driven instruction updates. Our work highlights the value of structured evaluation signals and establishes \model{} as a robust and generalizable tool for aligning video generation with human preferences.

% \section*{Limitations}

\bibliography{anthology,custom, aigve_score}

\begin{thebibliography}{62}
\providecommand{\natexlab}[1]{#1}

\bibitem[{Bai et~al.(2023)Bai, Bai, Yang, Wang, Tan, Wang, Lin, Zhou, and Zhou}]{bai_qwen-vl_2023}
Jinze Bai, Shuai Bai, Shusheng Yang, Shijie Wang, Sinan Tan, Peng Wang, Junyang Lin, Chang Zhou, and Jingren Zhou. 2023.
\newblock \href {http://arxiv.org/abs/2308.12966} {Qwen-{VL}: {A} {Versatile} {Vision}-{Language} {Model} for {Understanding}, {Localization}, {Text} {Reading}, and {Beyond}}.
\newblock \emph{arXiv preprint}.
\newblock ArXiv:2308.12966 [cs].

\bibitem[{Bai et~al.(2025)Bai, Chen, Liu, Wang, Ge, Song, Dang, Wang, Wang, Tang et~al.}]{bai2025qwen2}
Shuai Bai, Keqin Chen, Xuejing Liu, Jialin Wang, Wenbin Ge, Sibo Song, Kai Dang, Peng Wang, Shijie Wang, Jun Tang, and 1 others. 2025.
\newblock Qwen2. 5-vl technical report.
\newblock \emph{arXiv preprint arXiv:2502.13923}.

\bibitem[{Bansal et~al.(2024)Bansal, Lin, Xie, Zong, Yarom, Bitton, Jiang, Sun, Chang, and Grover}]{bansal2024videophy}
Hritik Bansal, Zongyu Lin, Tianyi Xie, Zeshun Zong, Michal Yarom, Yonatan Bitton, Chenfanfu Jiang, Yizhou Sun, Kai-Wei Chang, and Aditya Grover. 2024.
\newblock Videophy: Evaluating physical commonsense for video generation.
\newblock \emph{arXiv preprint arXiv:2406.03520}.

\bibitem[{Barratt and Sharma(2018)}]{barratt2018note}
Shane Barratt and Rishi Sharma. 2018.
\newblock A note on the inception score.
\newblock \emph{arXiv preprint arXiv:1801.01973}.

\bibitem[{Betker et~al.(2023)Betker, Goh, Jing, Brooks, Wang, Li, Ouyang, Zhuang, Lee, Guo et~al.}]{betker2023improving}
James Betker, Gabriel Goh, Li~Jing, Tim Brooks, Jianfeng Wang, Linjie Li, Long Ouyang, Juntang Zhuang, Joyce Lee, Yufei Guo, and 1 others. 2023.
\newblock Improving image generation with better captions (2023).
\newblock \emph{URL https://cdn. openai. com/papers/dall-e-3. pdf}.

\bibitem[{Caron et~al.(2021)Caron, Touvron, Misra, J{\'e}gou, Mairal, Bojanowski, and Joulin}]{caron2021emerging}
Mathilde Caron, Hugo Touvron, Ishan Misra, Herv{\'e} J{\'e}gou, Julien Mairal, Piotr Bojanowski, and Armand Joulin. 2021.
\newblock Emerging properties in self-supervised vision transformers.
\newblock In \emph{Proceedings of the IEEE/CVF international conference on computer vision}, pages 9650--9660.

\bibitem[{Chen et~al.(2022)Chen, Zhu, Li, Lu, Fan, and Wang}]{9452150}
Baoliang Chen, Lingyu Zhu, Guo Li, Fangbo Lu, Hongfei Fan, and Shiqi Wang. 2022.
\newblock \href {https://doi.org/10.1109/TCSVT.2021.3088505} {Learning generalized spatial-temporal deep feature representation for no-reference video quality assessment}.
\newblock \emph{IEEE Transactions on Circuits and Systems for Video Technology}, 32(4):1903--1916.

\bibitem[{Chen et~al.(2025)Chen, Sun, Tang, Li, and Chu}]{chen2025finger}
Rui Chen, Lei Sun, Jing Tang, Geng Li, and Xiangxiang Chu. 2025.
\newblock Finger: Content aware fine-grained evaluation with reasoning for ai-generated videos.
\newblock \emph{arXiv preprint arXiv:2504.10358}.

\bibitem[{Chen et~al.(2024)Chen, Sun, Tian, Jia, Zhang, Wang, Huang, Min, Zhai, and Zhang}]{chen2024gaiarethinkingactionquality}
Zijian Chen, Wei Sun, Yuan Tian, Jun Jia, Zicheng Zhang, Jiarui Wang, Ru~Huang, Xiongkuo Min, Guangtao Zhai, and Wenjun Zhang. 2024.
\newblock \href {https://arxiv.org/abs/2406.06087} {Gaia: Rethinking action quality assessment for ai-generated videos}.
\newblock \emph{Preprint}, arXiv:2406.06087.

\bibitem[{Cheng et~al.(2025)Cheng, Zhang, Zhang, Yang, Guan, Wu, Li, Zhang, Liu, Mai, Zeng, Wen, Jin, Wang, Zhou, Lu, Li, Huang, and Li}]{cheng2025simplevqamultimodalfactualityevaluation}
Xianfu Cheng, Wei Zhang, Shiwei Zhang, Jian Yang, Xiangyuan Guan, Xianjie Wu, Xiang Li, Ge~Zhang, Jiaheng Liu, Yuying Mai, Yutao Zeng, Zhoufutu Wen, Ke~Jin, Baorui Wang, Weixiao Zhou, Yunhong Lu, Tongliang Li, Wenhao Huang, and Zhoujun Li. 2025.
\newblock \href {https://arxiv.org/abs/2502.13059} {Simplevqa: Multimodal factuality evaluation for multimodal large language models}.
\newblock \emph{Preprint}, arXiv:2502.13059.

\bibitem[{Cho et~al.(2024)Cho, Hu, Baldridge, Garg, Anderson, Krishna, Bansal, Pont-Tuset, and Wang}]{Cho2024DSG}
Jaemin Cho, Yushi Hu, Jason Baldridge, Roopal Garg, Peter Anderson, Ranjay Krishna, Mohit Bansal, Jordi Pont-Tuset, and Su~Wang. 2024.
\newblock Davidsonian scene graph: Improving reliability in fine-grained evaluation for text-to-image generation.
\newblock In \emph{ICLR}.

\bibitem[{Ding et~al.(2021)Ding, Yang, Hong, Zheng, Zhou, Yin, Lin, Zou, Shao, Yang et~al.}]{ding2021cogview}
Ming Ding, Zhuoyi Yang, Wenyi Hong, Wendi Zheng, Chang Zhou, Da~Yin, Junyang Lin, Xu~Zou, Zhou Shao, Hongxia Yang, and 1 others. 2021.
\newblock Cogview: Mastering text-to-image generation via transformers.
\newblock \emph{Advances in neural information processing systems}, 34:19822--19835.

\bibitem[{He et~al.(2024{\natexlab{a}})He, Jiang, Zhang, Ku, Soni, Siu, Chen, Chandra, Jiang, Arulraj, Wang, Do, Ni, Lyu, Narsupalli, Fan, Lyu, Lin, and Chen}]{he-etal-2024-videoscore}
Xuan He, Dongfu Jiang, Ge~Zhang, Max Ku, Achint Soni, Sherman Siu, Haonan Chen, Abhranil Chandra, Ziyan Jiang, Aaran Arulraj, Kai Wang, Quy~Duc Do, Yuansheng Ni, Bohan Lyu, Yaswanth Narsupalli, Rongqi Fan, Zhiheng Lyu, Bill~Yuchen Lin, and Wenhu Chen. 2024{\natexlab{a}}.
\newblock \href {https://doi.org/10.18653/v1/2024.emnlp-main.127} {{V}ideo{S}core: Building automatic metrics to simulate fine-grained human feedback for video generation}.
\newblock In \emph{Proceedings of the 2024 Conference on Empirical Methods in Natural Language Processing}, pages 2105--2123, Miami, Florida, USA. Association for Computational Linguistics.

\bibitem[{He et~al.(2024{\natexlab{b}})He, Jiang, Zhang, Ku, Soni, Siu, Chen, Chandra, Jiang, Arulraj, Wang, Do, Ni, Lyu, Narsupalli, Fan, Lyu, Lin, and Chen}]{he2024videoscore}
Xuan He, Dongfu Jiang, Ge~Zhang, Max Ku, Achint Soni, Sherman Siu, Haonan Chen, Abhranil Chandra, Ziyan Jiang, Aaran Arulraj, Kai Wang, Quy~Duc Do, Yuansheng Ni, Bohan Lyu, Yaswanth Narsupalli, Rongqi Fan, Zhiheng Lyu, Yuchen Lin, and Wenhu Chen. 2024{\natexlab{b}}.
\newblock \href {https://arxiv.org/abs/2406.15252} {Videoscore: Building automatic metrics to simulate fine-grained human feedback for video generation}.
\newblock \emph{ArXiv}, abs/2406.15252.

\bibitem[{Hessel et~al.(2021)Hessel, Holtzman, Forbes, Le~Bras, and Choi}]{hessel-etal-2021-clipscore}
Jack Hessel, Ari Holtzman, Maxwell Forbes, Ronan Le~Bras, and Yejin Choi. 2021.
\newblock \href {https://doi.org/10.18653/v1/2021.emnlp-main.595} {{CLIPS}core: A reference-free evaluation metric for image captioning}.
\newblock In \emph{Proceedings of the 2021 Conference on Empirical Methods in Natural Language Processing}, pages 7514--7528, Online and Punta Cana, Dominican Republic. Association for Computational Linguistics.

\bibitem[{Ho et~al.(2022)Ho, Salimans, Gritsenko, Chan, Norouzi, and Fleet}]{ho2022videodiffusionmodels}
Jonathan Ho, Tim Salimans, Alexey Gritsenko, William Chan, Mohammad Norouzi, and David~J. Fleet. 2022.
\newblock \href {https://arxiv.org/abs/2204.03458} {Video diffusion models}.
\newblock \emph{Preprint}, arXiv:2204.03458.

\bibitem[{Hu et~al.(2023)Hu, Liu, Kasai, Wang, Ostendorf, Krishna, and Smith}]{hu2023tifa}
Yushi Hu, Benlin Liu, Jungo Kasai, Yizhong Wang, Mari Ostendorf, Ranjay Krishna, and Noah~A Smith. 2023.
\newblock \href {https://arxiv.org/abs/2303.11897} {Tifa: Accurate and interpretable text-to-image faithfulness evaluation with question answering}.
\newblock \emph{Preprint}, arXiv:2303.11897.

\bibitem[{Huang et~al.(2023)Huang, He, Yu, Zhang, Si, Jiang, Zhang, Wu, Jin, Chanpaisit, Wang, Chen, Wang, Lin, Qiao, and Liu}]{huang_vbench_2023}
Ziqi Huang, Yinan He, Jiashuo Yu, Fan Zhang, Chenyang Si, Yuming Jiang, Yuanhan Zhang, Tianxing Wu, Qingyang Jin, Nattapol Chanpaisit, Yaohui Wang, Xinyuan Chen, Limin Wang, Dahua Lin, Yu~Qiao, and Ziwei Liu. 2023.
\newblock \href {https://doi.org/10.48550/arXiv.2311.17982} {{VBench}: {Comprehensive} {Benchmark} {Suite} for {Video} {Generative} {Models}}.
\newblock \emph{arXiv preprint}.
\newblock ArXiv:2311.17982 [cs].

\bibitem[{Jin et~al.(2025)Jin, Sun, Li, Xu, Xu, Jiang, Zhuang, Huang, Song, Mu, and Lin}]{jin2024pyramidal}
Yang Jin, Zhicheng Sun, Ningyuan Li, Kun Xu, Kun Xu, Hao Jiang, Nan Zhuang, Quzhe Huang, Yang Song, Yadong Mu, and Zhouchen Lin. 2025.
\newblock Pyramidal flow matching for efficient video generative modeling.
\newblock In \emph{Proceedings of the International Conference on Learning Representations (ICLR)}.

\bibitem[{Kirstain et~al.(2023)Kirstain, Polyak, Singer, Matiana, Penna, and Levy}]{Kirstain2023PickaPicAO}
Yuval Kirstain, Adam Polyak, Uriel Singer, Shahbuland Matiana, Joe Penna, and Omer Levy. 2023.
\newblock Pick-a-pic: An open dataset of user preferences for text-to-image generation.

\bibitem[{Kong et~al.(2024)Kong, Tian, Zhang, Min, Dai, Zhou, Xiong, Li, Wu, Zhang, Wu, Lin, Yuan, Long, Wang, Wang, Li, Huang, Yang, Tan, Wang, Song, Bai, Wu, Xue, Wang, Wang, Liu, Li, Li, Wang, Yu, Deng, Li, Chen, Cui, Peng, Yu, He, Xu, Zhou, Xu, Tao, Lu, Liu, Zhou, Wang, Yang, Wang, Liu, Jiang, and Zhong}]{HunyuanVideo}
Weijie Kong, Qi~Tian, Zijian Zhang, Rox Min, Zuozhuo Dai, Jin Zhou, Jiangfeng Xiong, Xin Li, Bo~Wu, Jianwei Zhang, Kathrina Wu, Qin Lin, Junkun Yuan, Yanxin Long, Aladdin Wang, Andong Wang, Changlin Li, Duojun Huang, Fang Yang, and 33 others. 2024.
\newblock \href {https://arxiv.org/abs/2412.03603} {Hunyuanvideo: A systematic framework for large video generative models}.
\newblock \emph{arXiv preprint arXiv:2412.03603}.

\bibitem[{Kou et~al.(2024{\natexlab{a}})Kou, Liu, Zhang, Li, Wu, Min, Zhai, and Liu}]{kou_subjective-aligned_2024-1}
Tengchuan Kou, Xiaohong Liu, Zicheng Zhang, Chunyi Li, Haoning Wu, Xiongkuo Min, Guangtao Zhai, and Ning Liu. 2024{\natexlab{a}}.
\newblock \href {https://doi.org/10.48550/arXiv.2403.11956} {Subjective-{Aligned} {Dataset} and {Metric} for {Text}-to-{Video} {Quality} {Assessment}}.
\newblock \emph{arXiv preprint}.
\newblock ArXiv:2403.11956 [cs].

\bibitem[{Kou et~al.(2024{\natexlab{b}})Kou, Liu, Zhang, Li, Wu, Min, Zhai, and Liu}]{kou2024subjective}
Tengchuan Kou, Xiaohong Liu, Zicheng Zhang, Chunyi Li, Haoning Wu, Xiongkuo Min, Guangtao Zhai, and Ning Liu. 2024{\natexlab{b}}.
\newblock Subjective-aligned dataset and metric for text-to-video quality assessment.
\newblock In \emph{Proceedings of the 32nd ACM International Conference on Multimedia}, pages 7793--7802.

\bibitem[{Ku et~al.(2023)Ku, Jiang, Wei, Yue, and Chen}]{ku_viescore_2023}
Max Ku, Dongfu Jiang, Cong Wei, Xiang Yue, and Wenhu Chen. 2023.
\newblock \href {https://doi.org/10.48550/arXiv.2312.14867} {{VIEScore}: {Towards} {Explainable} {Metrics} for {Conditional} {Image} {Synthesis} {Evaluation}}.
\newblock \emph{arXiv preprint}.
\newblock ArXiv:2312.14867 [cs].

\bibitem[{Laurençon et~al.(2023)Laurençon, Saulnier, Tronchon, Bekman, Singh, Lozhkov, Wang, Karamcheti, Rush, Kiela, Cord, and Sanh}]{laurençon2023obelicsopenwebscalefiltered}
Hugo Laurençon, Lucile Saulnier, Léo Tronchon, Stas Bekman, Amanpreet Singh, Anton Lozhkov, Thomas Wang, Siddharth Karamcheti, Alexander~M. Rush, Douwe Kiela, Matthieu Cord, and Victor Sanh. 2023.
\newblock \href {https://arxiv.org/abs/2306.16527} {Obelics: An open web-scale filtered dataset of interleaved image-text documents}.
\newblock \emph{Preprint}, arXiv:2306.16527.

\bibitem[{Laurençon et~al.(2024)Laurençon, Tronchon, Cord, and Sanh}]{laurençon2024mattersbuildingvisionlanguagemodels}
Hugo Laurençon, Léo Tronchon, Matthieu Cord, and Victor Sanh. 2024.
\newblock \href {https://arxiv.org/abs/2405.02246} {What matters when building vision-language models?}
\newblock \emph{Preprint}, arXiv:2405.02246.

\bibitem[{Lee et~al.(2024)Lee, Yoon, Cho, and Bansal}]{lee2024videorepair}
Daeun Lee, Jaehong Yoon, Jaemin Cho, and Mohit Bansal. 2024.
\newblock Videorepair: Improving text-to-video generation via misalignment evaluation and localized refinement.
\newblock \emph{arXiv preprint arXiv:2411.15115}.

\bibitem[{Li et~al.(2024)Li, Lin, Pathak, Li, Fei, Wu, Ling, Xia, Zhang, Neubig, and Ramanan}]{li2024genaibenchevaluatingimprovingcompositional}
Baiqi Li, Zhiqiu Lin, Deepak Pathak, Jiayao Li, Yixin Fei, Kewen Wu, Tiffany Ling, Xide Xia, Pengchuan Zhang, Graham Neubig, and Deva Ramanan. 2024.
\newblock \href {https://arxiv.org/abs/2406.13743} {Genai-bench: Evaluating and improving compositional text-to-visual generation}.
\newblock \emph{Preprint}, arXiv:2406.13743.

\bibitem[{Li et~al.(2022)Li, Li, Xiong, and Hoi}]{li2022blipbootstrappinglanguageimagepretraining}
Junnan Li, Dongxu Li, Caiming Xiong, and Steven Hoi. 2022.
\newblock \href {https://arxiv.org/abs/2201.12086} {Blip: Bootstrapping language-image pre-training for unified vision-language understanding and generation}.
\newblock \emph{Preprint}, arXiv:2201.12086.

\bibitem[{Lin(2004)}]{lin-2004-rouge}
Chin-Yew Lin. 2004.
\newblock \href {https://aclanthology.org/W04-1013/} {{ROUGE}: A package for automatic evaluation of summaries}.
\newblock In \emph{Text Summarization Branches Out}, pages 74--81, Barcelona, Spain. Association for Computational Linguistics.

\bibitem[{Lin et~al.(2024)Lin, Cho, Zala, and Bansal}]{lin2024ctrladapterefficientversatileframework}
Han Lin, Jaemin Cho, Abhay Zala, and Mohit Bansal. 2024.
\newblock \href {https://arxiv.org/abs/2404.09967} {Ctrl-adapter: An efficient and versatile framework for adapting diverse controls to any diffusion model}.
\newblock \emph{Preprint}, arXiv:2404.09967.

\bibitem[{Liu et~al.(2024{\natexlab{a}})Liu, Li, Li, and Lee}]{liu2024improvedbaselinesvisualinstruction}
Haotian Liu, Chunyuan Li, Yuheng Li, and Yong~Jae Lee. 2024{\natexlab{a}}.
\newblock \href {https://arxiv.org/abs/2310.03744} {Improved baselines with visual instruction tuning}.
\newblock \emph{Preprint}, arXiv:2310.03744.

\bibitem[{Liu et~al.(2024{\natexlab{b}})Liu, Li, Li, Li, Zhang, Shen, and Lee}]{liu2024llavanext}
Haotian Liu, Chunyuan Li, Yuheng Li, Bo~Li, Yuanhan Zhang, Sheng Shen, and Yong~Jae Lee. 2024{\natexlab{b}}.
\newblock \href {https://llava-vl.github.io/blog/2024-01-30-llava-next/} {Llava-next: Improved reasoning, ocr, and world knowledge}.

\bibitem[{Liu et~al.(2024{\natexlab{c}})Liu, Xiang, Li, Wang, Li, Liu, Zhang, Ye, and Zhang}]{liu2024surveyaigve}
Xiao Liu, Xinhao Xiang, Zizhong Li, Yongheng Wang, Zhuoheng Li, Zhuosheng Liu, Weidi Zhang, Weiqi Ye, and Jiawei Zhang. 2024{\natexlab{c}}.
\newblock \href {https://arxiv.org/abs/2410.19884} {A survey of ai-generated video evaluation}.
\newblock \emph{arXiv preprint arXiv:2410.19884}.

\bibitem[{Liu et~al.(2023{\natexlab{a}})Liu, Iter, Xu, Wang, Xu, and Zhu}]{liu2023gevalnlgevaluationusing}
Yang Liu, Dan Iter, Yichong Xu, Shuohang Wang, Ruochen Xu, and Chenguang Zhu. 2023{\natexlab{a}}.
\newblock \href {https://arxiv.org/abs/2303.16634} {G-eval: Nlg evaluation using gpt-4 with better human alignment}.
\newblock \emph{Preprint}, arXiv:2303.16634.

\bibitem[{Liu et~al.(2023{\natexlab{b}})Liu, Cun, Liu, Wang, Zhang, Chen, Liu, Zeng, Chan, and Shan}]{liu_evalcrafter_2023}
Yaofang Liu, Xiaodong Cun, Xuebo Liu, Xintao Wang, Yong Zhang, Haoxin Chen, Yang Liu, Tieyong Zeng, Raymond Chan, and Ying Shan. 2023{\natexlab{b}}.
\newblock \href {https://doi.org/10.48550/arXiv.2310.11440} {{EvalCrafter}: {Benchmarking} and {Evaluating} {Large} {Video} {Generation} {Models}}.
\newblock \emph{arXiv preprint}.
\newblock ArXiv:2310.11440 [cs].

\bibitem[{Ma et~al.(2022)Ma, Xu, Sun, Yan, Zhang, and Ji}]{ma2022x}
Yiwei Ma, Guohai Xu, Xiaoshuai Sun, Ming Yan, Ji~Zhang, and Rongrong Ji. 2022.
\newblock X-clip: End-to-end multi-grained contrastive learning for video-text retrieval.
\newblock In \emph{Proceedings of the 30th ACM international conference on multimedia}, pages 638--647.

\bibitem[{Miao et~al.(2024)Miao, Zhu, Dong, Yu, Zhu, and Gao}]{miao2024t2vsafety}
Yibo Miao, Yifan Zhu, Yinpeng Dong, Lijia Yu, Jun Zhu, and Xiao-Shan Gao. 2024.
\newblock \href {https://arxiv.org/abs/2407.05965} {T2vsafetybench: Evaluating the safety of text-to-video generative models}.
\newblock \emph{Preprint}, arXiv:2407.05965.

\bibitem[{Mittal et~al.(2012)Mittal, Moorthy, and Bovik}]{mittal2012no}
Anish Mittal, Anush~Krishna Moorthy, and Alan~Conrad Bovik. 2012.
\newblock \href {https://doi.org/10.1109/TIP.2012.2214050} {No-reference image quality assessment in the spatial domain}.
\newblock \emph{IEEE Transactions on Image Processing}, 21(12):4695--4708.

\bibitem[{OpenAI(2024{\natexlab{a}})}]{gpt4o}
OpenAI. 2024{\natexlab{a}}.
\newblock Gpt-4o.
\newblock \url{https://openai.com/index/hello-gpt-4o/}.

\bibitem[{OpenAI(2024{\natexlab{b}})}]{sora2024}
OpenAI. 2024{\natexlab{b}}.
\newblock Sora.
\newblock \url{https://openai.com/index/sora/}.

\bibitem[{OpenAI(2025)}]{gpt41}
OpenAI. 2025.
\newblock Gpt-4.1.
\newblock \url{https://openai.com/index/gpt-4-1/}.

\bibitem[{Ramesh et~al.(2021)Ramesh, Pavlov, Goh, Gray, Voss, Radford, Chen, and Sutskever}]{ramesh2021zero}
Aditya Ramesh, Mikhail Pavlov, Gabriel Goh, Scott Gray, Chelsea Voss, Alec Radford, Mark Chen, and Ilya Sutskever. 2021.
\newblock Zero-shot text-to-image generation.
\newblock In \emph{International conference on machine learning}, pages 8821--8831. Pmlr.

\bibitem[{Singer et~al.(2022)Singer, Polyak, Hayes, Yin, An, Zhang, Hu, Yang, Ashual, Gafni, Parikh, Gupta, and Taigman}]{singer2022makeavideotexttovideogenerationtextvideo}
Uriel Singer, Adam Polyak, Thomas Hayes, Xi~Yin, Jie An, Songyang Zhang, Qiyuan Hu, Harry Yang, Oron Ashual, Oran Gafni, Devi Parikh, Sonal Gupta, and Yaniv Taigman. 2022.
\newblock \href {https://arxiv.org/abs/2209.14792} {Make-a-video: Text-to-video generation without text-video data}.
\newblock \emph{Preprint}, arXiv:2209.14792.

\bibitem[{Team(2024{\natexlab{a}})}]{geminiteam2024gemini15unlockingmultimodal}
Gemini Team. 2024{\natexlab{a}}.
\newblock \href {https://arxiv.org/abs/2403.05530} {Gemini 1.5: Unlocking multimodal understanding across millions of tokens of context}.
\newblock \emph{Preprint}, arXiv:2403.05530.

\bibitem[{Team(2024{\natexlab{b}})}]{genmo2024mochi}
Genmo Team. 2024{\natexlab{b}}.
\newblock Mochi 1.
\newblock \url{https://github.com/genmoai/models}.

\bibitem[{Unterthiner et~al.(2019)Unterthiner, van Steenkiste, Kurach, Marinier, Michalski, and Gelly}]{fvd}
Thomas Unterthiner, Sjoerd van Steenkiste, Karol Kurach, Raphael Marinier, Marcin Michalski, and Sylvain Gelly. 2019.
\newblock \href {https://arxiv.org/abs/1812.01717} {Towards accurate generative models of video: A new metric and challenges}.
\newblock \emph{Preprint}, arXiv:1812.01717.

\bibitem[{Venkatanath et~al.(2015)Venkatanath, Praneeth, Bh, Channappayya, and Medasani}]{venkatanath2015blind}
Narasimhan Venkatanath, D~Praneeth, Maruthi~Chandrasekhar Bh, Sumohana~S Channappayya, and Swarup~S Medasani. 2015.
\newblock Blind image quality evaluation using perception based features.
\newblock In \emph{2015 twenty first national conference on communications (NCC)}, pages 1--6. IEEE.

\bibitem[{Verma and Leddo()}]{vermacomparing}
Shreyas Verma and John Leddo.
\newblock Comparing the effectiveness between human-generated videos and ai-generated videos on learning.

\bibitem[{Wang et~al.(2025)Wang, Yang, Tuo, He, Zhu, Fu, and Liu}]{wang2025swap}
Wenjing Wang, Huan Yang, Zixi Tuo, Huiguo He, Junchen Zhu, Jianlong Fu, and Jiaying Liu. 2025.
\newblock Swap attention in spatiotemporal diffusions for text-to-video generation.
\newblock \emph{International Journal of Computer Vision}, pages 1--19.

\bibitem[{Wang et~al.(2004)Wang, Bovik, Sheikh, and Simoncelli}]{1284395}
Zhou Wang, A.C. Bovik, H.R. Sheikh, and E.P. Simoncelli. 2004.
\newblock \href {https://doi.org/10.1109/TIP.2003.819861} {Image quality assessment: from error visibility to structural similarity}.
\newblock \emph{IEEE Transactions on Image Processing}, 13(4):600--612.

\bibitem[{Wu et~al.(2023{\natexlab{a}})Wu, Liao, Wang, Chen, Hou, Sun, Yan, and Lin}]{BVQI}
Haoning Wu, Liang Liao, Annan Wang, Chaofeng Chen, Jingwen Hou, Wenxiu Sun, Qiong Yan, and Weisi Lin. 2023{\natexlab{a}}.
\newblock \href {https://arxiv.org/abs/2304.14672} {Towards robust text-prompted semantic criterion for in-the-wild video quality assessment}.
\newblock \emph{Preprint}, arXiv:2304.14672.

\bibitem[{Wu et~al.(2023{\natexlab{b}})Wu, Zhang, Liao, Chen, Hou, Wang, Sun, Yan, and Lin}]{wu_exploring_2023-1}
Haoning Wu, Erli Zhang, Liang Liao, Chaofeng Chen, Jingwen Hou, Annan Wang, Wenxiu Sun, Qiong Yan, and Weisi Lin. 2023{\natexlab{b}}.
\newblock \href {http://arxiv.org/abs/2211.04894} {Exploring {Video} {Quality} {Assessment} on {User} {Generated} {Contents} from {Aesthetic} and {Technical} {Perspectives}}.
\newblock \emph{arXiv preprint}.
\newblock ArXiv:2211.04894 [cs, eess].

\bibitem[{Xiang et~al.(2025)Xiang, Liu, Li, Liu, and Zhang}]{xiang2025aigvetoolaigeneratedvideoevaluation}
Xinhao Xiang, Xiao Liu, Zizhong Li, Zhuosheng Liu, and Jiawei Zhang. 2025.
\newblock \href {https://arxiv.org/abs/2503.14064} {Aigve-tool: Ai-generated video evaluation toolkit with multifaceted benchmark}.
\newblock \emph{arXiv preprint arXiv:2503.14064}.

\bibitem[{Yang et~al.(2024)Yang, Teng, Zheng, Ding, Huang, Xu, Yang, Hong, Zhang, Feng et~al.}]{yang2024cogvideox}
Zhuoyi Yang, Jiayan Teng, Wendi Zheng, Ming Ding, Shiyu Huang, Jiazheng Xu, Yuanming Yang, Wenyi Hong, Xiaohan Zhang, Guanyu Feng, and 1 others. 2024.
\newblock Cogvideox: Text-to-video diffusion models with an expert transformer.
\newblock \emph{arXiv preprint arXiv:2408.06072}.

\bibitem[{Zhang et~al.(2025)Zhang, Li, Cheng, Hu, Yuan, Chen, Leng, Jiang, Zhang, Li, Jin, Zhang, Wang, Bing, and Zhao}]{damonlpsg2025videollama3}
Boqiang Zhang, Kehan Li, Zesen Cheng, Zhiqiang Hu, Yuqian Yuan, Guanzheng Chen, Sicong Leng, Yuming Jiang, Hang Zhang, Xin Li, Peng Jin, Wenqi Zhang, Fan Wang, Lidong Bing, and Deli Zhao. 2025.
\newblock \href {https://arxiv.org/abs/2501.13106} {Videollama 3: Frontier multimodal foundation models for image and video understanding}.
\newblock \emph{Preprint}, arXiv:2501.13106.

\bibitem[{Zhang et~al.(2023)Zhang, Wu, Liu, Zhao, Ran, Gu, Gao, and Shou}]{zhang2023show1marryingpixellatent}
David~Junhao Zhang, Jay~Zhangjie Wu, Jia-Wei Liu, Rui Zhao, Lingmin Ran, Yuchao Gu, Difei Gao, and Mike~Zheng Shou. 2023.
\newblock \href {https://arxiv.org/abs/2309.15818} {Show-1: Marrying pixel and latent diffusion models for text-to-video generation}.
\newblock \emph{Preprint}, arXiv:2309.15818.

\bibitem[{Zhang et~al.(2024{\natexlab{a}})Zhang, Tian, Huang, Qiao, and Liu}]{zhang2024evaluation}
Fan Zhang, Shulin Tian, Ziqi Huang, Yu~Qiao, and Ziwei Liu. 2024{\natexlab{a}}.
\newblock Evaluation agent: Efficient and promptable evaluation framework for visual generative models.
\newblock \emph{arXiv preprint arXiv:2412.09645}.

\bibitem[{Zhang et~al.(2020)Zhang, Kishore, Wu, Weinberger, and Artzi}]{zhang2020bertscoreevaluatingtextgeneration}
Tianyi Zhang, Varsha Kishore, Felix Wu, Kilian~Q. Weinberger, and Yoav Artzi. 2020.
\newblock \href {https://arxiv.org/abs/1904.09675} {Bertscore: Evaluating text generation with bert}.
\newblock \emph{Preprint}, arXiv:1904.09675.

\bibitem[{Zhang et~al.(2024{\natexlab{b}})Zhang, Zhang, Goh, and Sun}]{Zhang2024The}
Xinyi Zhang, Renyu Zhang, K.~Goh, and Chenshuo Sun. 2024{\natexlab{b}}.
\newblock \href {https://doi.org/10.48550/arXiv.2412.18337} {The value of ai-generated metadata for ugc platforms: Evidence from a large-scale field experiment}.
\newblock \emph{ArXiv}, abs/2412.18337.

\bibitem[{Zhong et~al.(2022)Zhong, Liu, Yin, Mao, Jiao, Liu, Zhu, Ji, and Han}]{zhong2022unifiedmultidimensionalevaluatortext}
Ming Zhong, Yang Liu, Da~Yin, Yuning Mao, Yizhu Jiao, Pengfei Liu, Chenguang Zhu, Heng Ji, and Jiawei Han. 2022.
\newblock \href {https://arxiv.org/abs/2210.07197} {Towards a unified multi-dimensional evaluator for text generation}.
\newblock \emph{Preprint}, arXiv:2210.07197.

\bibitem[{Zhou et~al.(2024)Zhou, Liu, Dong, Kou, Gao, Zhang, Li, Wu, and Zhai}]{Light_VQA_plus}
Xunchu Zhou, Xiaohong Liu, Yunlong Dong, Tengchuan Kou, Yixuan Gao, Zicheng Zhang, Chunyi Li, Haoning Wu, and Guangtao Zhai. 2024.
\newblock \href {https://arxiv.org/abs/2405.03333} {Light-vqa+: A video quality assessment model for exposure correction with vision-language guidance}.
\newblock \emph{Preprint}, arXiv:2405.03333.

\end{thebibliography}

\appendix
\newpage
\onecolumn

\section{Implementation Details}
\label{app:imple}
We implement \model{} based on the Qwen2.5-VL-7B~\cite{bai2025qwen2} architecture, initializing from the official checkpoint\footnote{\url{https://huggingface.co/Qwen/Qwen2.5-VL-7B-Instruct}}. During finetuning, the vision encoder is kept frozen, while all remaining parameters are updated. We set the weight for score and comment to 50. Training is conducted on two NVIDIA A6000 GPUs using a learning rate of 1e-5. We train the model for 3 epochs and apply early stopping based on validation loss. Optimization is performed using AdamW, with a linear learning rate scheduler and 5\% warmup steps. The full training process takes approximately 24 hours.

\section{Ablation Study}
\label{app:ablation}

\begin{table*}[h]
\centering
\resizebox{\textwidth}{!}{
\begin{tabular}{l|ccccccccc|c}
\toprule
\textbf{Method} & \textbf{Technical} & \textbf{Dynamic} & \textbf{Consistency} & \textbf{Physics} & \textbf{Element\_Pre} & \textbf{Element\_Qu} & \textbf{Act\_Pre} & \textbf{Act\_Qu} & \textbf{Overall} & \textbf{G-Eval} \\
\midrule
\model{} & \textbf{40.60} & \textbf{57.31} & \textbf{61.49} & \textbf{64.36} & \textbf{40.32} & \textbf{40.81} & \textbf{44.31} & \textbf{60.71} & \textbf{59.88} & \textbf{3.42} \\
-weighted loss & 33.15 & 48.33 & 41.92 & 44.81 & 30.10 & 38.21 & 44.10 & 55.71 & 48.32 & 2.97 \\
-dyna sample & 37.96 & 50.76 & 47.52 & 40.41 & 35.36 & 33.08 & 38.12 & 53.24 & 53.99 & 3.31 \\
Qwen2.5-VL & 8.77 & 4.00 & 1.24 & -6.01 & 9.19 & 10.19 & 18.74 & 0.72 & 9.59 & 2.37\\
\bottomrule
\end{tabular}
}
\caption{Ablation study of our finetuning strategy.}
\label{tab:ablation}
\end{table*}

The ablation results in Table~\ref{tab:ablation} clearly demonstrate the effectiveness of both components in the proposed finetuning strategy—token-wise weighted loss and dynamic frame sampling. Removing the weighted loss leads to a sharp performance drop across nearly all evaluation aspects, most notably in Consistency (-19.57) and Physics (-19.55), suggesting that emphasizing score and comment tokens is crucial for aligning with human judgment. Similarly, removing dynamic sampling substantially hurts aspects sensitive to temporal change, such as Dynamic and Action Quality, confirming its role in capturing meaningful motion cues. Compared to the pretrained Qwen2.5-VL, the full model achieves large gains across the board (e.g., +51.7 in Consistency, +60.4 in Physics), showing that joint modeling of structured outputs with targeted loss design and adaptive input selection is key to high-quality, aspect-aware video evaluation.

\section{Details of \data{}}
\label{app:data_details}
\subsection{Evaluation Aspect Description}
\begin{table*}[h!]
    \centering
    \small
    \renewcommand{\arraystretch}{1.2} % Increases row height
    \vspace{\baselineskip} 
    \scalebox{0.95}{ % Slightly scales the table
    % \begin{tabular}{l|p{13cm}}
    \begin{tabularx}{\linewidth}{l|X}
        \toprule
        \textbf{Metric} & \textbf{Description} \\
        \midrule
        \textbf{Technical Quality} & Assesses the technical aspects of the video, including whether the resolution is sufficient for object recognition, whether the colors are natural, and whether there is an absence of noise or artifacts. \\
        \rowcolor[HTML]{F2F2F2}  \textbf{Dynamic} & Measures the extent of pixel changes throughout the video, focusing on significant object or camera movements and changes in environmental factors such as daylight, weather, or seasons. \\
        \textbf{Consistency} & Evaluates whether objects in the video maintain consistent properties, avoiding glitches, flickering, or unexpected changes. \\
        \rowcolor[HTML]{F2F2F2} \textbf{Physics} & Determines if the scene adheres to physical laws, ensuring that object behaviors and interactions are realistic and aligned with real-world physics. \\
        \textbf{Element Presence} & Checks if all objects mentioned in the instructions are present in the video. The score is based on the proportion of objects that are correctly included. \\
        \rowcolor[HTML]{F2F2F2} \textbf{Element Quality} & Assesses the realism and fidelity of objects in the video, awarding higher scores for detailed, natural, and visually appealing appearances. \\
        \textbf{Action/Interaction Presence} & Evaluates whether all actions and interactions described in the instructions are accurately represented in the video. \\
        \rowcolor[HTML]{F2F2F2} \textbf{Action/Interaction Quality} & Measures the naturalness and smoothness of actions and interactions, with higher scores for those that are realistic, lifelike, and seamlessly integrated into the scene. \\
        \textbf{Overall} & Reflects the comprehensive quality of the video based on all metrics, allowing raters to incorporate their subjective preferences into the evaluation. \\
        \bottomrule
    % \end{tabular}
    \end{tabularx}
    }
    \caption{Metrics for Video Generation Evaluation}
    \label{tab:metrics}
\end{table*}

\newpage

\subsection{\data{} Comment Length Analysis}
\begin{figure}[h]
    \centering
    \includegraphics[width=0.5\linewidth]{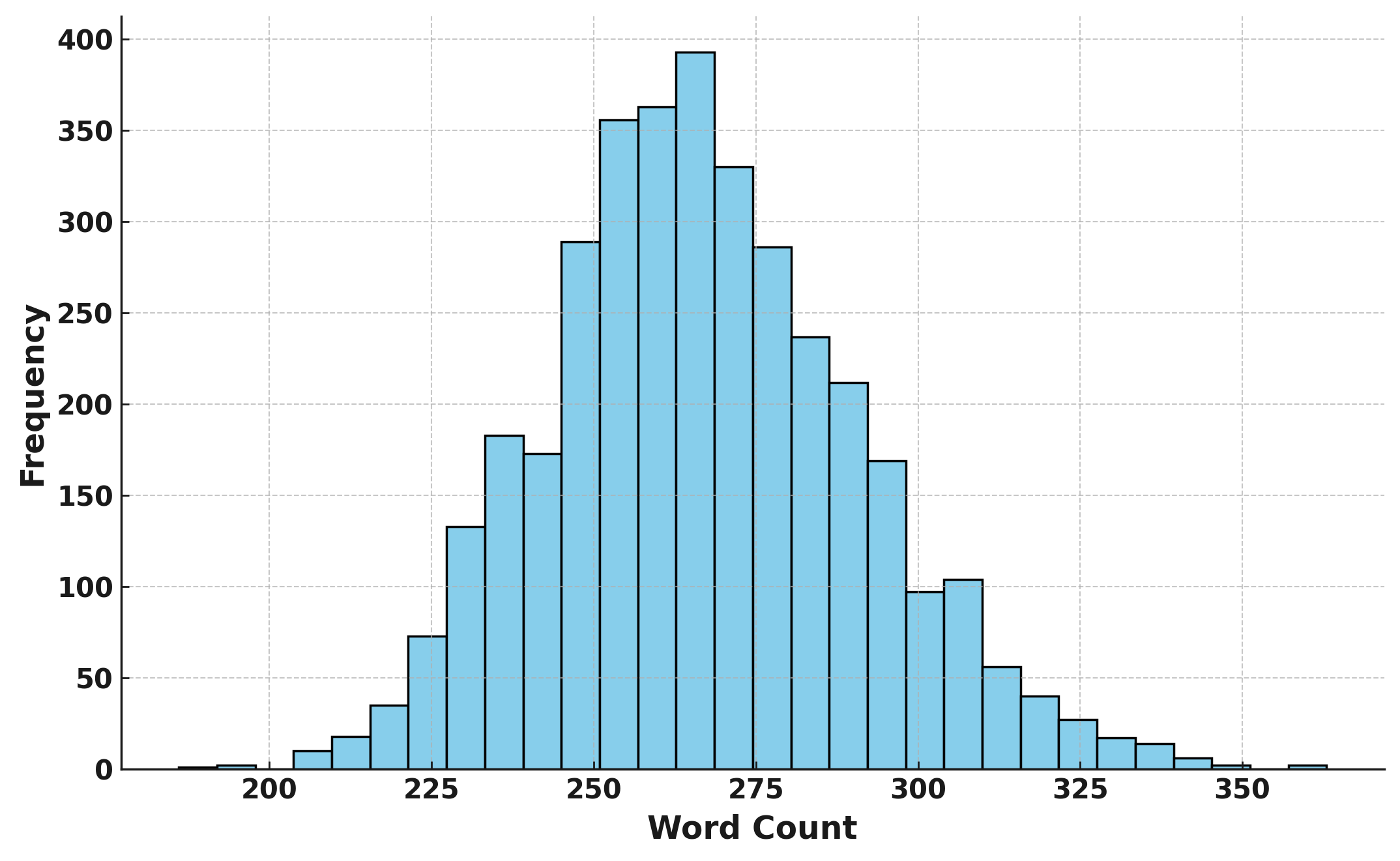}
    \caption{Comment Word Number Distribution of \data{}.}
    \label{fig:comment_length_distribution}
\end{figure}

The distribution of combined comment word counts indicates that the revised comments are notably rich and thorough, averaging around 267 words per entry. This high word count reflects a strong emphasis on providing contextually rich, multi-aspect evaluations.
\newpage
\section{More Case Studies}
\label{app:case_studies}
\begin{figure*}[!h]
    \centering
    \begin{subfigure}[t]{0.98\linewidth}
        \centering
        \includegraphics[width=\linewidth]{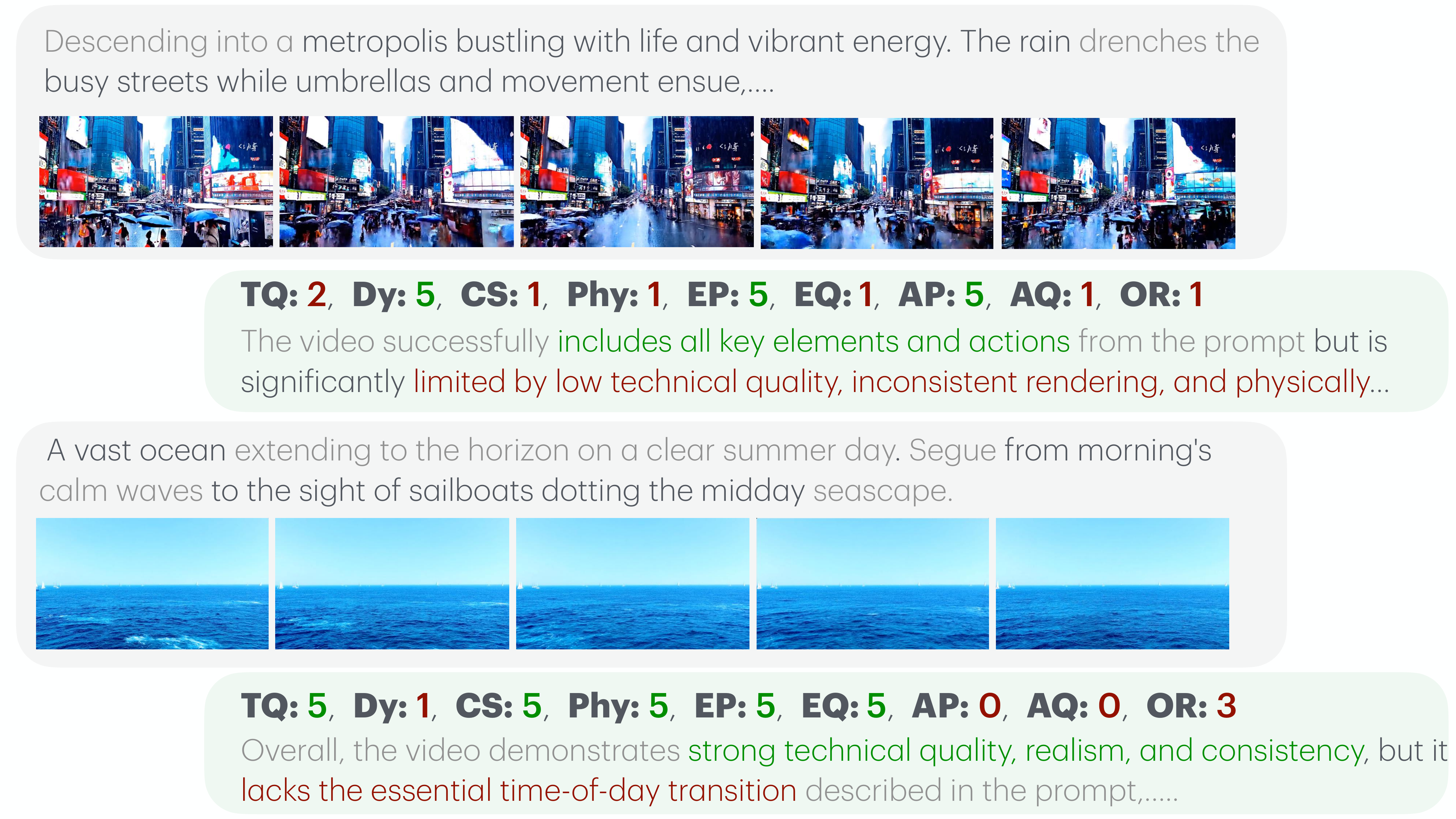}
    \end{subfigure}
    
    \vspace{0.5em} % adjust vertical space between them
    
    \begin{subfigure}[t]{0.98\linewidth}
        \centering
        \includegraphics[width=\linewidth]{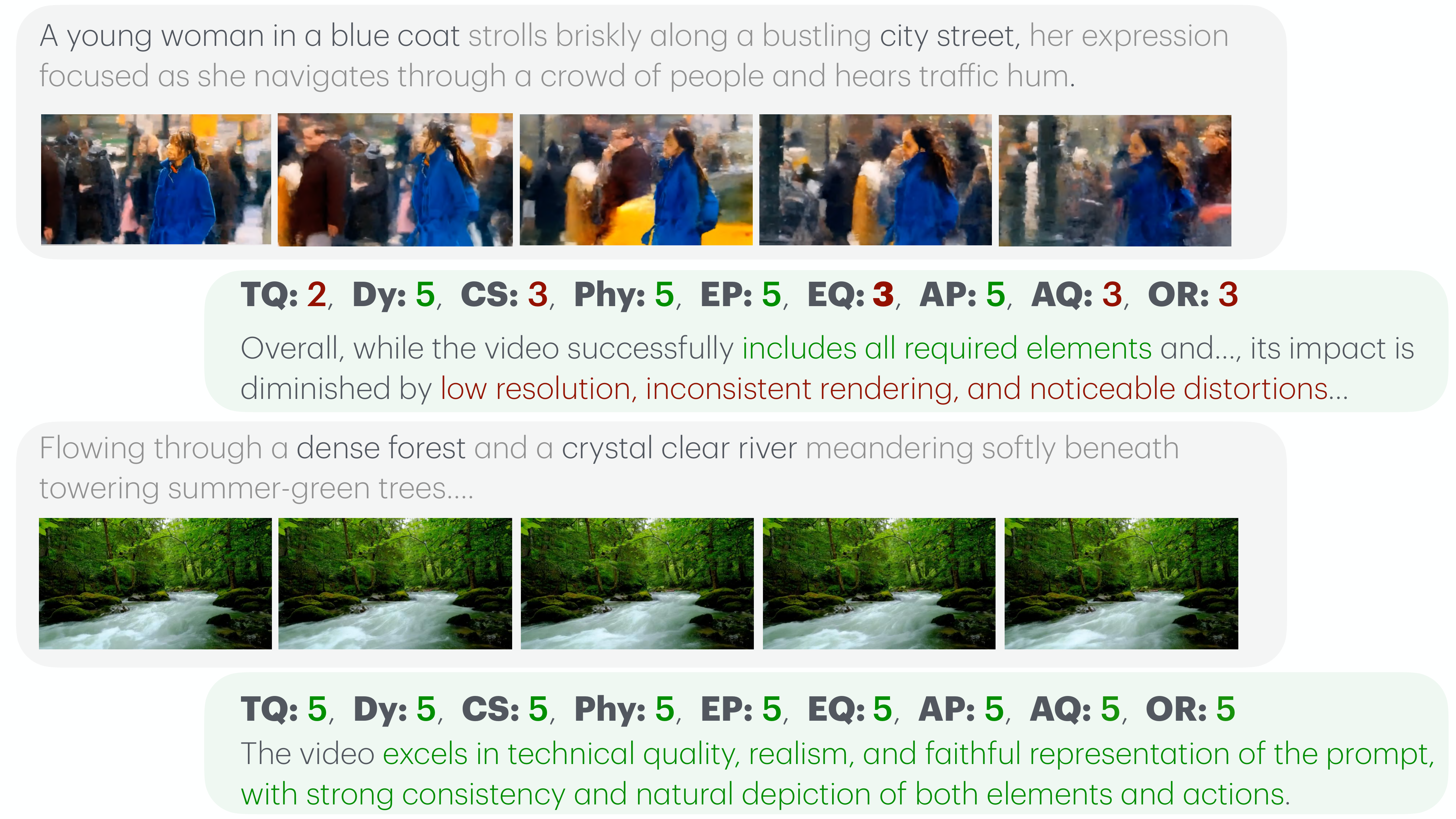}
    \end{subfigure}
    
    \caption{Case Study of \model{}.}
    \label{fig:case_study_app}
\end{figure*}

\end{document}